%% file: main.tex
\documentclass{article}

\usepackage{arxiv}
\usepackage[utf8]{inputenc} 
\usepackage[T1]{fontenc}    
\usepackage[hyphens]{url}            
\usepackage{booktabs}       
\usepackage{amsfonts}       
\usepackage{amssymb}
\usepackage{amsmath, mathtools, amsthm} 
\usepackage{thmtools}
\usepackage{float}
\usepackage{nicefrac}       
\usepackage{microtype}      
\usepackage{lipsum}
\usepackage{graphicx}
\usepackage{caption}
\usepackage{subcaption}
\usepackage{multirow, array} 
\usepackage{todonotes}
\usepackage{footnote} 
\usepackage[normalem]{ulem}
\usepackage{cancel} 
\usepackage[bottom]{footmisc} 
\usepackage{arydshln} 

\usepackage{color}
\usepackage{actuarialangle} 
\usepackage{dsfont} 

\usepackage[
    backend=biber,
    style=numeric,
    citestyle=numeric-comp,
    sorting = nyt,
    maxbibnames = 99
]{biblatex}
\DeclareNameAlias{author}{last-first}

\renewbibmacro{in:}{%
  \ifentrytype{article}{}{%
  \printtext{\bibstring{in}\intitlepunct}}}

\renewbibmacro*{volume+number+eid}{\printfield{volume}\printfield{number}\printfield{eid}}
\DeclareFieldFormat[article]{number}{(#1)}

\DeclareFieldFormat{pages}{#1}

\DeclareFieldFormat[article,inbook,incollection,inproceedings,patent,thesis,unpublished]{title}{{#1\isdot}}

\usepackage{tikz,pgfplots,pgf}
\usepackage{tkz-euclide}
\usetikzlibrary{matrix,shapes,positioning}
\usetikzlibrary{arrows,backgrounds}
\usepgflibrary{shapes.multipart}

\usepackage[many]{tcolorbox}
\newtcolorbox{cross}{blank,breakable,parbox=false,
  overlay={\draw[red,line width=2pt] (interior.south west)--(interior.north east);
    \draw[red,line width=2pt] (interior.north west)--(interior.south east);}}

\usepackage{geometry}
\declaretheoremstyle[notefont=\normalfont]{normalhead}

\declaretheorem[style=normalhead]{proof*}
\declaretheorem[style=normalhead]{definition}

\declaretheorem[style=normalhead]{remark}

\DeclareCaptionFormat{cont}{#1 (cont.)#2#3\par}

\title{Neural calibration of hidden inhomogeneous Markov chains \\ \large Information decompression in life insurance}

\author{
    Mark Kiermayer \\ 
  Department of Natural Science\\
  University of Applied Sciences Ruhr West\\
 \texttt{mark.kiermayer@hs-ruhrwest.de} \\ 
   \And
    Christian Weiß\\
  Department of Natural Sciences\\
  University of Applied Sciences Ruhr West\\
  \texttt{christian.weiss@hs-ruhrwest.de} \\
}

\begin{document}
\maketitle
\raggedbottom

\begin{abstract}
    Markov chains play a key role in a vast number of areas, including life insurance mathematics. Standard actuarial quantities as the premium value can be interpreted as compressed, lossy information about the underlying Markov process. We introduce a method to reconstruct the underlying Markov chain given collective information of a portfolio of contracts. Our neural architecture explainably characterizes the process by explicitly providing one-step transition probabilities. Further, we provide an intrinsic, economic model validation to inspect the quality of the information decompression. Lastly, our methodology is successfully tested for a realistic data set of German term life insurance contracts.
\end{abstract}


\input{Sections/Intro}

\input{Sections/Objective}

\input{Sections/Data}

\input{Sections/Model}
\input{Sections/Numerical_Results}

\input{Sections/Conclusion}

\section*{Acknowledgements}
The authors want to thank msg-life central europe gmbh for providing the data set and in particular Dr. Stefan Nörtemann and Volker Dietz for numerous discussions and invaluable feedback. Further, the authors are grateful to the 'Ministeriums für Kultur und Wissenschaft des Landes Nordrhein-Westfalen' for supporting their research by the grant 'FH BASIS 2019' (reference 1908fhb005). All numerical experiments presented in this work were conducted on a server funded by the respective grant. Last, the first named author would like to thank the Saint Petersburg Electrotechnical University "LETI" for the hospitality during his stay at the faculty of computer science and technology.

\printbibliography

\input{Sections/Appendix}

\end{document}

%% file: Sections/Intro.tex
\section{Introduction} \label{section:introduction}

\begin{figure}[htb]
    \centering
    \input{Sections/Sketch_workflow}
    \caption{Generic workflow of actuarial computations with non-observable transition probabilities.}
    \label{figure:workflow}
\end{figure}
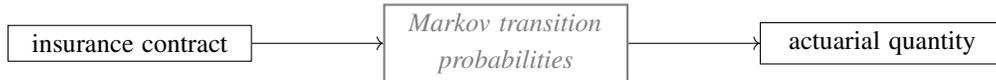

Markov chains present a widely applicable concept for modeling processes with a rich theory, see e.g. \cite{Borovkov.2013, Durrett.2010, Klenke.2014}, which allows to relax the common assumption of identically distributed and independent data in sequential settings. 
In practice, we find numerous examples and areas of application including, but not limited to, medicine \cite{Asanjarani.2021, Ren.2015}, finance \cite{Ferreira.2020, Cai.1994}, strategic planning processes \cite{Verbeken.2021, Sutton.2018, Levine.2018,Ding.2018} and natural language processing \cite{Blunsom.2004, Khan.2016, Zhu.2009}. In the particular case of life insurance, Markov processes and the related transition probabilities between states lie at the core of the business for managing and pricing risk, see e.g. \cite{Fuehrer:2010, Gerber.2013}. Therefore, they require special attention. Transition probabilities are traditionally estimated homogeneously from observable state histories of policyholders and are then displayed as a table, such as the German DAV 2008T table \cite{DAV:2008T}, or as a continuous, parametric model, see e.g. \cite{dickson:2009}.  
Recent work in \cite{Deprez.2017, Perla.2021, Richman.2019, Richman.2021} aims to enhance the quality of parametric mortality models, such as the Lee-Carter model, 
by using models from machine learning and deep learning to calibrate them. This also contributes to a profound, multidisciplinary branch of research that utilizes neural networks as a modelling approach in Markov settings, see e.g. \cite{Awiszus.2018, Qu.2019, Khurana.2020, Liu.2020}.\\
In the present work, we look at a rather general setting. Some actuarial quantity is observed, e.g. a premium or policy values, that is computed as a function of a corresponding insurance contract and potentially inhomogeneous Markov transition probabilities. We illustrate the computational flow in Figure \ref{figure:workflow}. For a given portfolio of insurance contracts and the corresponding actuarial quantities, our objective is to reconstruct the underlying Markov process. Similarly to literature listed above, we utilize the class of neural networks to optimize our objective. However, in contrast to \cite{Deprez.2017, Perla.2021, Richman.2019, Richman.2021}, we retrieve hidden Markov assumptions that an insurance company originally imposed when setting up the contracts. This information allows us to identify different profiles of policyholders in a portfolio, e.g. smokers and non-smokers, each potentially which varying risk surcharges. A second practical use case is a migration of contracts to a new IT system, when information may be lost or automated backtesting is required. 

\paragraph{Contribution.}
We introduce a novel approach for fitting transition probabilities in a setting where no state histories of individuals are observed. The methodology is fully explainable, in the sense that individual, one-step transition probabilities can be inspected and analyzed. The transition probabilities are generally inhomogeneous and Markovian. In the special case of a single, non-terminal state, our approach can also reconstruct a semi-Markov process. In order to economically confirm the quality of the calibrated transition probabilities, we present an intrinsic model validation. Further, our method is successfully implemented for a data set of realistically computed, German term life insurance contracts, which were generated by msg life central europe gmbh and their administrative system for life insurance.    

\paragraph{Outline.} In Section \ref{section:framework}, we combine relevant information from both the theory of Markov processes and the respective actuarial context and formalize our objective. Section \ref{section:data} summarizes the real-world data, as well as relevant assumption for the numeric analysis. The neural architecture and our intrinsic model evaluation are then presented in Section \ref{section:model}, followed by the numerical results in Section \ref{section:num_experiments}. Lastly, we conclude with a summary and an outlook for further research in Section \ref{section:conclusion}.



%% file: Sections/Sketch_workflow.tex
\begin{tikzpicture}
    \node[draw, align=center, text width = 3cm] (one) at (0,0) {insurance contract};
    \node[draw,align=center, text width = 3cm, thick, gray] (two) at (5,0) {\textit{Markov transition probabilities}};
    \node[draw, align=center, text width = 3cm] (three) at (10,0) {actuarial quantity};

    \draw [->] (one) to (two);
    \draw [->] (two) to (three);
\end{tikzpicture}

%% file: Sections/Objective.tex
\section{General framework} \label{section:framework}
\begin{figure}[htb]
        \centering
        \begin{subfigure}{.45\textwidth}
            \centering
            \input{Sections/MC_example_1}
        \end{subfigure}
        \begin{subfigure}{.45\textwidth}
            \centering
            \input{Sections/MC_example_2}
        \end{subfigure}
        \caption{Examples for common Markov type settings with two (left) and three (right) states.  }
        \label{fig:markov:example}
    \end{figure}
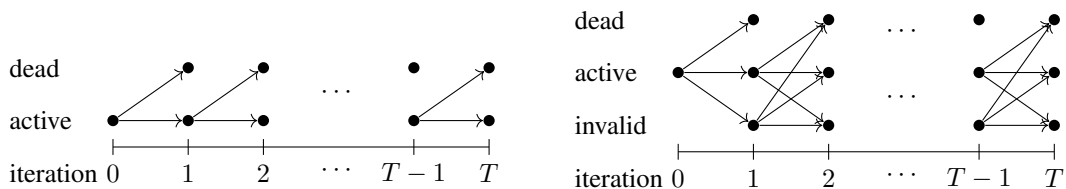

    Figure \ref{fig:markov:example} illustrates two common examples in life insurance, where a policyholder evolves over time and can change or maintain its state of being active, invalid or dead. Each combination of time and state is associated with a cash flow, indicated by black nodes. Each policyholder will follow one of the paths displayed in Figure \ref{fig:markov:example}. To determine the value of the underlying policy at time $0$, an insurance company needs to quantify for all cash flows their probability of being realized. For this purpose Markov chains are commonly used, see e.g. \cite{dickson:2009, Fuehrer:2010}.

\paragraph{Markov chains in life insurance.}
    Let $(\Omega, \mathcal{A}, \mathbb{P})$ be a probability space and the random vector $C:\Omega\rightarrow\mathbb{R}^{n}$ describe an insurance contract of a fixed type, e.g. an endowment or a term life insurance. A realization $c\in C(\Omega)$ thus presents a numeric encoding of a contract and may include policyholder-related features, such as age, gender or smoker status, as well as policy-related features, such as the maximum duration of the contract, the frequency of premium payments or a sum insured. Further, let 
    $$X(c)=\lbrace X^{(k)}(c)\vert~X^{(k)}(c): \Omega\rightarrow \mathcal{S}\rbrace_{k\in\mathbb{N}_0}$$ 
    denote a discrete stochastic process which maps a contract $c\in C(\Omega)$ to the countable state space $\mathcal{S}$. \\
    The state space $\mathcal{S}$ in life insurance traditionally includes the policyholder being 'active', 'invalid' or 'dead'. In practice, the dynamics of $X(c)$ are affected only by certain components in the vector $c$, commonly including age and gender and excluding e.g. the premium amount. For illustrative wording, we use the state of the policyholder interchangeably with the state of the contract, although the state is generally also affected by policy-related features such as the maximum duration.  \\
    
    Transition probabilities between states $i,j\in\mathcal{S}$ are a key component for actuarial computations. They allow to quantify the probability of ending up in an arbitrary state $j$ at time $k$ and, thereby, to compute the expected value of all cash flows. In the most general form, such transition probabilities can depend on the complete history of the path and take the form of
    \begin{align*} 
        \mathbb{P}\left(X^{(k+1)}(c)= j\vert X^{(k)}(c)=i, X^{(k-1)}(c)=s_{k-1},\ldots, X^{(0)}(c)=s_0\right)
    \end{align*}
    for states $i,j,s_{k-1},\ldots, s_0 \in\mathcal{S}$. \\
    As an unlimited dependence on the past is both impractical and computationally challenging, we follow the common practice and assume $X^{(k)}(c)$ to be a \textit{Markov chain}, see \cite{Borovkov.2013, Klenke.2014, Durrett.2010}. Hence, we can limit the dependence on the history to the latest observation. We write transition probabilities $p_{ij}^{(k)}(c)$ by
    \begin{align*}
        p_{ij}^{(k)}(c) &:=\mathbb{P}\left(X^{(k+1)}(c)= j\vert X^{(k)}(c)=i\right)
        \intertext{and introduce the matrix $\pi^{(k)}(c)$ of transition probabilities by}
        \pi^{(k)}(c) &:= \left(p_{ij}^{(k)}(c)\right)_{i,j\in\mathcal{S}}.
    \end{align*}
    Note that $p_{ij}^{(k)}(c)$ can depend on the iteration $k$. Hence, $X^{(k)}(c)$ is an \textit{inhomogeneous Markov chain}, see e.g. \cite{Borovkov.2013, Klenke.2014}. \\
    
    
    Eventually, to describe a path of arbitrary length we need to combine one step transition probabilities to multiple steps. Since $X(c)$ is a Markov chain, it satisfies the \textit{Chapman-Kolmogorov equation}. Therefore multi-step transition probabilities $M^{(n,k)}(c)\in[0,1]^{\vert \mathcal{S}\vert\times\vert \mathcal{S}\vert}$ between time $n$ and $n+k$, $n,k\in\mathbb{N}_0$, are characterized by
    \begin{equation}
         \begin{aligned} \label{eq:markov:multistep}
            M^{(n,k)}(c) &:=\left(\mathbb{P}\left(X^{(n+k)}(c)= j\vert X^{(n)}(c)=i \right)\right)_{i,j\in\mathcal{S}}
             \\
             &~= \prod_{l =0}^{k-1}\pi^{(n+l)}(c).
        \end{aligned}   
    \end{equation}
    
    With \eqref{eq:markov:multistep}, we can now state the probability of realizing cash flows at arbitrary points in time, as e.g. in the introductory example in Figure \ref{fig:markov:example}. This will be a key component to our objective, where we calibrate $\pi^{(k)}(c)$ explicitly in multi-period settings.

\paragraph{Actuarial context.} 
    Next, we provide an overview of a common actuarial principle. It will establish a connection between transition probabilities and cash flows, which we will exploit in our objective. For more detail on actuarial mathematics we refer the reader to \cite{dickson:2009, Fuehrer:2010, Gerber.2013}. \\
    A general evaluation of an insurance product is equivalent to computing the \textit{actuarial present value} (APV) of all potential cash flows the design of the product entails. Let $c\in C(\Omega)$ be a contract and $s_0$ be a fixed, initial state of the process $X(c)$. The state $s_0$ commonly corresponds to the policyholder being alive at the start of the contract. Further, let $CF_{ij}^{(k)}(c)$ denote cash flows for to being in state $j\in\mathcal{S}$ at iteration $k\in\mathbb{N}$ after having been in $i\in\mathcal{S}$ at time $k-1$. For initial cash flows, which are irrespective of any state transition, we use the analogous notation of $CF_{ij}^{(0)}(c)$ and set $CF_{ij}^{(0)}(c)=0$ if $i,j\neq s_0$.  Cash flows are then recorded $m\in\mathbb{N}$ times per year and are subject to a constant annual discount factor $v\in(0,1)$. Ultimately, the APV of $c$ can be computed by
    \begin{equation}
        \begin{aligned}
            \text{APV}(c) &:= \sum_{i,j \in\mathcal{S}} CF_{ij}^{(0)}+\sum_{k\in\mathbb{N},i,j\in\mathcal{S}} \mathbb{P}\left(X^{(k)}(c)= j, X^{(k-1)}(c)= i\vert X^{(0)}(c)=s_0\right)  CF_{ij}^{(k)}(c)~ v^{\tfrac{k}{m}} \\
            \label{eq:apv:discounted}
            & := \sum_{i,j \in\mathcal{S}}y_{ij}^{(0)}(c)+\sum_{k,i,jk\in\mathbb{N},i,j\in\mathcal{S}} \mathbb{P}\left(X^{(k)}(c)= j, X^{(k-1)}(c)= i\vert X^{(0)}(c)=s_0\right) y_{ij}^{(k)}(c),
        \end{aligned}    
    \end{equation}
    where $y_{ij}^{(k)}(c):=CF_{ij}^{(k)}(c)~ v^{\tfrac{k}{m}}$ is the discounted cash flow for policy $c$ at iteration $k$ and in state $j$. 
    \begin{definition} \label{def:apv:consistent}
        We call a contract $c\in C(\Omega)$ \textit{APV-consistent} if APV$(c)=0$ is satisfied.
    \end{definition}
    
    Our nomenclature in Definition \ref{def:apv:consistent} is motivated by the general concept of a market-consistent evaluation of assets and liabilities, see e.g. \cite{Wuthrich.2010, Wuthrich.2011}. The concept of APV-consistency is commonly used to determine e.g. policy values, alias reserves, or calibrate premium values, which both can be interpreted as components of cash flows. In Section \ref{section:data}, we will see practical examples of cash flows $CF_{ij}^{(k)}$ for term life insurance contracts. 

\paragraph{Objective.} 
    Finally, we combine \eqref{eq:markov:multistep} and \eqref{eq:apv:discounted}, in order to recapture the full dynamics $(\pi^{(k)}(c))_{k\in\mathbb{N}_0}$ of the state process $X(c)$ for an arbitrary APV-consistent contract $c$. \\ 
    Let $\hat{\pi}: C(\Omega)\times\mathbb{N}_0 \rightarrow [0,1]^{\vert \mathcal{S}\vert \times\vert \mathcal{S}\vert}$ denote an estimator that maps pairs $(c,k)$, with contract $c$ at iteration $k$, to a matrix $\hat{\pi}^{(k)}(c)$ of one-step transition probabilities of $X(c)$ with initial state $X^{(0)}(c):=s_0$. The matrix of $k$-step transition probabilities is then estimated by $\hat{M}^{(0,k)}(c):=\prod_{l=0}^{k-1} \hat{\pi}^{(l)}(c)\in [0,1]^{\vert \mathcal{S}\vert \times\vert \mathcal{S}\vert}$, recall \eqref{eq:markov:multistep}. Further, we introduce the class $\mathcal{F}:=\lbrace \hat{\pi}: C(\Omega)\times\mathbb{N}_0 \rightarrow [0,1]^{\vert \mathcal{S}\vert \times\vert \mathcal{S}\vert}\rbrace$ of all potential candidates $\hat{\pi}$. Next, let a contract $c\in C(\Omega)$ be APV-consistent, $\ell: \mathbb{R}\rightarrow\mathbb{R}_{\geq 0}$ be a norm and $\psi:\mathcal{F}\times C(\Omega)\times \mathbb{R}^{\vert \mathcal{S}\vert\times\vert \mathcal{S}\vert \times \mathbb{N}_0}\rightarrow \mathbb{R}$ the function, which implements the economic computation in \eqref{eq:apv:discounted}. We define $\psi$ by
    \begin{align}\label{eq:psi}
        \psi\left(\hat{\pi},c,y(c)\right) &:=  \sum_{k\in\mathbb{N}_0,i,j\in\mathcal{S}} \hat{M}^{(0,k-1)}_{s_0, i}(c)~\hat{\pi}^{(k-1)}_{ij}(c)~ y_{ij}^{(k)}(c),
    \end{align}
    where we set $\hat{\pi}^{(-1)}_{ij}(c):=\mathds{1}_{\lbrace s_0,s_0\rbrace}(i,j)$ and $y(c):=\left(y_{ij}^{(k)}(c)\right)_{i,j\in\mathcal{S},k\in\mathbb{N}_0}$. \\
    
    The optimal estimator $\pi^{\star}\in\mathcal{F}$ is then characterized  by
    \begin{align}\label{eq:objective}
        \pi^{\star} &:=  \arg\!\min_{\hat{\pi}\in\mathcal{F}} \mathbb{E}\left[ \ell \circ \psi\left(\hat{\pi},C,y(C)\right)\right].
    \end{align}
    Note that by employing the loss function $\ell\circ \psi$, we match quantities of different type and scale, i.e. the unbounded, real-valued random vector $C$, probability matrices $\hat{\pi}^{(k)}(C)$ and the tensor $y(C)$.
    
    \begin{remark}
        \begin{itemize}
            \item[1.] One should think of \eqref{eq:objective} as a decompression of information, i.e. transition probabilities $\pi^{(k)}(c)$, contained in contracts $c\in C(\Omega)$ and the APV-consistent cash flows. In practice, this information can naturally at most be reconstructed on areas of the feature space which are populated by observations. 
            \item[2.] Transition probabilities $\hat{\pi}^{(k)}(c)$ depend not only on the iteration $k\in\mathbb{N}_0$, but also features of contract $c$ such as e.g. age, gender, smoker status. This makes the objective a challenging, high-dimensional setting for optimization.
            \item[3.] For the application in Section \ref{section:model}, the major advantageous of using a composite loss function $\ell\circ \psi$ is that via $\psi$ we export some of the complexity from the network architecture to the loss function. Further, in $y_{ij}^{(k)}(c)$ we can simultaneously process distinct assumptions for individual contracts, although later analysis will not utilize that flexibility.
            \item[4.] Quantities $y_{ij}^{(k)}(c)$ are generally not available. Hence, in the numeric analysis we will utilize prior information on the type of contract and work with estimates.
            \item[5.] In practice, the setup of a contract commonly includes a risk buffer, such as a multiplicative $34\%$ buffer on mortality rates as in the German mortality table DAV 2008T in \cite{DAV:2008T}. In this case, our objective will directly retrieve so called \textit{first-order transition rates}, which include the risk buffer. Conversely, this means that if we were to consider contracts with a non-APV-consistent, ex-post premium loading as e.g. in an Italian data set in \cite{Aleandri.2017}, we would have to first reverse this loading. Otherwise it will be factored into transition probabilities ${\pi^{\star}}^{(k)}(c)$.
        \end{itemize}
    \end{remark}

%% file: Sections/MC_example_1.tex
\begin{tikzpicture}
    
    \tikzstyle{knot}=[circle, fill=black, minimum size = 0.1pt];
    
    \draw (0,0) -- (5,0);
    
    \foreach \x in {0,1,2,4,5}
      \draw (\x cm,3pt) -- (\x cm,-3pt);
    \foreach \x/\y in {0,1,2,4/T-1,5/T}
        \node[centered] at (\x, -10pt) {$\y$};
    \foreach \x in {0,1,2,4,5}   
        \node [circle,fill,inner sep=1.5pt] (alive-\x) at (\x, 10pt){};
     \foreach \x [count=\y] in {1,2,4,5}    
        \node [circle,fill,inner sep=1.5pt] (dead-\x) at (\x, 30pt){};
    \foreach \x/\y in {1/0,2/1,5/4}
        \draw [->] (alive-\y) to (alive-\x);
    \foreach \x/\y in {1/0,2/1,5/4}
        \draw [->] (alive-\y) to (dead-\x);

    \node[text width = 3cm, right] at (-1.5, -10pt) {iteration};
    \node[text width = 3cm, right] at (-1.5, 10pt) {active};
    \node[text width = 3cm, right] at (-1.5, 30pt) {dead};
    \node[text width = 3cm, right] at (-1.5, 50pt) {}; 
    \draw (3,0) node[below=3pt] {$ \dotsi $} node[above=15pt] {$ \dotsi $};

\end{tikzpicture}

%% file: Sections/MC_example_2.tex
\begin{tikzpicture}
    
    \tikzstyle{knot}=[circle, fill=black, minimum size = 0.1pt];
    
    \draw (0,0) -- (5,0);
    
    \foreach \x in {0,1,2,4,5}
      \draw (\x cm,3pt) -- (\x cm,-3pt);
    \foreach \x/\y in {0,1,2,4/T-1,5/T}
        \node[centered] at (\x, -10pt) {$\y$};
    \foreach \x in {0,1,2,4,5}   
        \node [circle,fill,inner sep=1.5pt] (alive-\x) at (\x, 30pt){};
     \foreach \x [count=\y] in {1,2,4,5}    
        \node [circle,fill,inner sep=1.5pt] (dead-\x) at (\x, 50pt){};
    \foreach \x [count=\y] in {1,2,4,5}    
        \node [circle,fill,inner sep=1.5pt] (invalid-\x) at (\x, 10pt){};
    \foreach \x/\y in {1/0,2/1,5/4}
        \draw [->] (alive-\y) to (alive-\x);
    \foreach \x/\y in {1/0,2/1,5/4}
        \draw [->] (alive-\y) to (dead-\x);
    \foreach \x/\y in {1/0,2/1,5/4}
        \draw [->] (alive-\y) to (invalid-\x);
    \foreach \x/\y in {2/1,5/4}
        \draw [->] (invalid-\y) to (invalid-\x);
    \foreach \x/\y in {2/1,5/4}
        \draw [->] (invalid-\y) to (dead-\x);
    \foreach \x/\y in {2/1,5/4}
        \draw [->] (invalid-\y) to (alive-\x);

    \node[text width = 3cm, right] at (-1.5, -10pt) {iteration};
    \node[text width = 3cm, right] at (-1.5, 10pt) {invalid};
    \node[text width = 3cm, right] at (-1.5, 30pt) {active};
    \node[text width = 3cm, right] at (-1.5, 50pt) {dead};
    \draw (3,0) node[below=3pt] {$ \dotsi $} node[above=15pt] {$ \dotsi $} node[above=40pt] {$ \dotsi $};
\end{tikzpicture}

%% file: Sections/Data.tex
\section{Data} \label{section:data}
All data are generated and provided by msg life central europe gmbh, a leading insurance software provider. The data set consists of $N=10,000$ German term life insurance contracts from the years 2015-2016, which were selected from the 'msg.Life Factory', a real-world administration system for insurance policies. In this section we describe the nature of the data set and how to preprocess it to solve our objective. 

\paragraph{Term life insurance.}
A term life insurance  follows a plain logic, which is introduced in more detail e.g. in \cite{Fuehrer:2010, dickson:2009, Gerber.2013}. A policyholder (resp. their spouses) will receive a sum insured $S\in\mathbb{R}$, if they die within a fixed period of $n$ years. In return for that financial prospect the policyholder has to pay a premium amount $P$ for $t\leq n$ years in a predetermined frequency\footnote{In general, paying a lump sum up-front ($m=0)$ is admissible. We exclude this option, as it is not present in our data set. Although $m=0$  is actually a simpler version of our task, it requires some notational attention, since e.g. a linear scaling from annual premium values $P$ to sub-annual values $P/m$ as in the upcoming equations \eqref{eq:cf:active} and \eqref{eq:cf:dead} is inadmissible. } $m\in\mathbb{N}$, where e.g. $m=1$ corresponds to annual payments and $m=4$ quarterly payments. The insurance company, on top of potentially having to pay the sum insured, incurs expenses for acquisition and administration of the contract.  All quantities mentioned above are linked since the premium amount $P$ is set in a way such the expected present value of all cash flows equals zero, including a risk buffer\footnote{In Germany the risk buffer is applied multiplicatively to the transition probabilities.}. \\ 
This approach for calibrating the premium is commonly known as the actuarial equivalence principle, see e.g. \cite{dickson:2009, Fuehrer:2010, Gerber.2013}. In our data we will observe the scalar premium value $P$ for each contract. The precise assumptions on how these values are obtained are unknown.

\paragraph{Portfolio.}
We now describe the nature of our data. For a clearer presentation, we follow the notation of the previous paragraph and refer to individual features by letters instead of indexing the vector of features $c$. Tables \ref{tab:data:quantitative} and \ref{tab:data:qualitative} show information on individual features of the data set $\mathcal{C}$ of term life insurance contracts $c\in\mathcal{C}$, with $\vert \mathcal{C}\vert = 10,000$. The data contains the 'year' and 'month' the contract was initiated, as well as the 'age' $a_0$ of the policyholder at that time. Further, we observe the maximum duration $n$ of the contract, the duration $t$ of premium payments. The data stems from the years 2015 and 2016 and cover a wide range of ages from 18 to 60 and duration $n$ and $t$ as short as one year and up to 48 and 46 years. Moreover, the values of the premium and sum insured can be as low as 1,000€ and 6€ p.a. and as high as 1,000,000€ and 200,000€ p.a. Overall, the statistics indicate the features 'year', 'month' 'age' and 'sum insured' to be approximately uniformly distributed, while 'n', 't' and the 'premium' are skewed to the right. This is visually confirmed by a plot of the marginal distributions, see Figure \ref{fig:appendix:data:marginal:dist} in the Appendix. 
Additionally, there are three qualitative features in the data, namely how frequently are premiums paid ('payment style'), the gender of the policyholder and their smoker status. As in Section \ref{section:framework}, we explicitly denote the 'payment style' by $m\in\mathbb{N}$ payments per year. In Table \ref{tab:data:qualitative} we report all levels and observe that they are roughly uniformly distributed for all qualitative features. 

\begin{table}[htb]
    \centering
    \begin{tabular}{l|ccccccc}
        {} &  year &  month &         $a_0$: age &         $n$ &         $t$ &   $S$: sum insured &        $P$: premium \\
        \midrule
        mean  &     2015.49 &         6.44 &     39.19 &     13.92 &      7.57 &  504876.88 &    9634.36 \\
        std   &        0.50 &         3.45 &     12.45 &     10.62 &      7.43 &  288506.79 &   14885.05 \\
        min   &     2015 &         1 &     18 &      1 &      1 &    1001.98 &       5.95 \\
        25\%   &     2015 &         3 &     29 &      5 &      2 &  250213.84 &    2231.38 \\
        75\%   &     2016 &         9 &     50 &     21 &     11 &  755203.49 &   10767.67 \\
        max   &     2016 &        12 &     60 &     48 &     46 &  999841.57 &  206828.02 \\
        \bottomrule
    \end{tabular}
    \caption{Statistics of quantitative features.}
    \label{tab:data:quantitative}
\end{table}

\begin{table}[htb]
    \centering
    \begin{tabular}{l|l|l}
                  $m$: payment style  &       $g$: gender &          smoker \\
        \midrule
            12: monthly (0.26) &  male (0.50) &  yes (0.49) \\
              4:{ } quarterly (0.25)   &  female (0.50) &  no (0.51)    \\
              2:{ } semi-annually (0.24)   &                     &                        \\
          1:{ } annually (0.25) &                     &                        \\
          \bottomrule
    \end{tabular}
    \caption{Levels and relative frequency of qualitative features.}
    \label{tab:data:qualitative}
\end{table}

\paragraph{Further assumptions.} We assume that contracts are APV-consistent. Further, in order to implement our objective \eqref{eq:objective}, we impose a structure for the cash flows of term life insurance contracts, resulting in estimated cash flows $\widehat{CF}_{ij}^{(k)}$ for a given iteration $k$ and state transition from $j$ to $i$. For the most part, the structure of cash flows is encoded in the contract features and the type of contract. In our two-state-model, we encode the state 'alive' by 0 and 'dead' by 1. Through the information in the contract $c$, inhomogeneous transition probabilities $p_{ij}^{(k)}(c)$ at iteration $k$ account for features such as the current age of the policyholder and the step width $m\in\mathbb{N}$, alias the payment style. In calendar time, at the $k$-th iteration of $(X^{(k)})_{k\in \mathbb{N}_0}$ a time of $\tfrac{k}{m}$ years has passed. We look at the cash flow from the perspective of the insurance company\footnote{The opposite perspective of the policyholder is equally valid, since in our objective we compose the economic computation $\psi$ with the norm $\ell$, recall Section \ref{section:framework}.}, where the annual premium $P$ is perceived as income. On the other side, the potential payment of the sum insured $S$ is a liability and additional expenses occur. Therefore, we employ the common $(\alpha, \beta, \gamma_1, \gamma_2)$ \textit{expense structure}, see e.g. \cite{Fuehrer:2010, Burkhart.2018}. These factors are related to the acquisition of the contract ($\alpha$), the collection of premiums ($\beta$) and administrative charges during the time of premium payments ($\gamma_1$) and thereafter ($\gamma_2$). We then define our estimates for the true cash flows $CF_{ij}^{(k)}(c)$ for $k\in\mathbb{N}_0$ by
\begin{align} 
        \label{eq:cf:active}
        \widehat{CF}_{00}^{(k)}(c) :=&~ \frac{P}{m}\mathds{1}_{\lbrace k/m<t\rbrace} - t\alpha P\mathds{1}_{\lbrace k=0\rbrace} - \beta \frac{P}{m}\mathds{1}_{\lbrace k/m<t\rbrace} - \gamma_1\frac{S}{m}\mathds{1}_{\lbrace k/m<t\rbrace} - \gamma_2\frac{S}{m}\mathds{1}_{\lbrace t\leq k/m<n\rbrace}, \\
        \label{eq:cf:dead}
        \widehat{CF}_{01}^{(k)}(c) :=&~ \left(\frac{P}{m}\mathds{1}_{\lbrace k/m<t\rbrace} - S  - \beta \frac{P}{m}\mathds{1}_{\lbrace k/m<t\rbrace} - \gamma_1\frac{S}{m}\mathds{1}_{\lbrace k/m<t\rbrace} - \gamma_2\frac{S}{m}\mathds{1}_{\lbrace t\leq k/m<n\rbrace}\right)\mathds{1}_{\lbrace k>0\rbrace},\\
    & ~\hspace{50pt} \widehat{CF}_{10}^{(k)}(c) := 0 \quad\quad \text{ and } \quad\quad \widehat{CF}_{11}^{(k)}(c) := 0.
\end{align}
Next, we introduce an annual \textit{discount factor} $\hat{v}\in(0,1]$. This allows us to respect the time component of iteration $k\in\mathbb{N}_0$ and express it as a present value at time 0. Therefore, we estimate the quantity $y\in\mathbb{R}^{2\times 2\times n/m+1}$ as in \eqref{eq:apv:discounted} by
\begin{align} \label{eq:cf:discounted}
    \hat{y}_{ij}^{(k)}(c) := \widehat{CF}_{ij}^{(k)}(c)~\hat{v}^{\tfrac{k}{m}}, \quad\quad i,j=1,2, ~k=0,1,\ldots,\frac{n}{m}.
\end{align}

Further, we form a realistic cost structure with estimated values
\begin{align*}
    (\alpha, \beta, \gamma_1, \gamma_2) &:= (0.025, 0.03, 0.001, 0.001)
\end{align*}
by consulting applied research in \cite{Burkhart.2018}. Since the maximum admissible actuarial interest rate in Germany for the years 2015 and 2016 is $1.25\%$, see \cite{DAV:Hoechstrechnungszins}, we set the discount factor $\hat{v}$ to
\begin{align*}
    \hat{v} &:= 1.0125^{-1}.
\end{align*}
Finally, we generate the semi-observable data set $\mathcal{D}:= \lbrace c, \hat{y}(c) \rbrace_{c\in\mathcal{C}}$.

\begin{remark}[Including prior information]
    \begin{itemize}
         \item[1.] The structure in \eqref{eq:cf:active} - \eqref{eq:cf:discounted} presents prior information which we introduce in our task. We require the existence of such information, but it can easily be adapted to other types, as e.g. endowment or whole life insurance, see e.g. \cite{dickson:2009, Fuehrer:2010}. Conceptually, quantities $\widehat{CF}_{ij}^{(k)}(c)$ resemble rewards in a reinforcement learning framework, see e.g. \cite{Sutton.2018}.
        \item[2.] Discounting in \eqref{eq:cf:discounted} implicitly assumes all components, i.e. summands, of a cash flow $\widehat{CF}_{ij}^{(k)}$ to occur at the beginning of each period $[k, k+1)$. If sufficiently reliable prior information is available, one can push the timing of individual components in $\widehat{CF}_{ij}^{(k)}$ to an arbitrary time $t\in[0,1]$ within the period by multiplying the respective summand(s) by $\hat{v}^{t/m}$. 
        \item[3.] All assumptions to form the unobservable, discounted cash flows $\hat{y}_{ij}(c)$ are per contract $c\in \mathcal{C}$. Hence, for different contracts $c$ one might employ e.g. different expense structures $(\alpha, \beta, \gamma_1, \gamma_2)$. In this work, however, we use the same assumptions for all contracts $c\in\mathcal{C}$.
    \end{itemize}
\end{remark}

\begin{remark}[Hyperparameters]
    \begin{itemize}
        \item[1.] The hyperparameters for constructing $\hat{y}(c)$ are crucial. Since in our objective \eqref{eq:objective} the true quantities $y(c)$ are fixed, any consistently wrong estimate, such as a too low discount factor $\hat{v}$, will be implicitly factored into the transition probabilities $\hat{\pi}^{(k)}(c)$.
        \item[2.] We can tune hyperparameters, such as $\alpha, \beta$ or $\hat{v}$, in a similar manner as we will tune the neural architecture in Section \ref{section:model}. However, this is computationally expensive as the data set $\mathcal{D}$ itself depends on these specific values.
    \end{itemize}
\end{remark}

\paragraph{Mortality tables.}
The DAV 2008T table \cite{DAV:2008T} presents a guideline for life insurance and homogeneous transition probabilities in Germany. The presented values all refer to annual transitions of inter-valued ages of policyholders. Although these values generally do not present an optimal solution for our objective \eqref{eq:objective}, they are still useful as a reasonable baseline. We will address this more detailed in Section \ref{section:model}.\\
We denote the homogeneous probabilities in the table for the current age $a$ and the payment style $m$ by $\Tilde{\pi}(a,m,g)$ for a fixed $g\in\lbrace \text{male, female} \rbrace$. Values $\Tilde{\pi}_{01}(a,1,g)$ can then be read directly from the table. For sub-annual payment styles $m>1,~m\in\mathbb{N}$, we assume death events to be uniformly distributed\footnote{For practical reasons, this assumption could be hard-encoded in the later network architecture. However, to restrict the number of assumptions, we will refrain from doing so in Section \ref{section:model}.} over the year, compare e.g. \cite{Fuehrer:2010, dickson:2009}. Hence, for all $a,m\in\mathbb{N}$ and $g\in\lbrace \text{male, female} \rbrace$ it holds 
\begin{align*}
    \Tilde{\pi}_{01}(a,m,g) &:=\frac{1}{m}\Tilde{\pi}_{01}(a,1,g), \\
    \Tilde{\pi}_{00}(a,m,g) &:= 1 -\Tilde{\pi}_{01}(a,m,g),
\end{align*}    
and naturally
\begin{align*}
    \Tilde{\pi}_{11}(a,m,g) := 1 \quad\quad \text{ and } \quad \quad \Tilde{\pi}_{10}(a,m,g) := 0.
\end{align*}
We then take the range of ages $A:= \lbrace 0,1,\ldots,121\rbrace$ in the DAV 2008T table and the collection of payment frequencies $M:=\lbrace 1,2,4,12 \rbrace$ in the data $\mathcal{C}$ and form the data set $\mathcal{D}_{DAV}(g):= \lbrace (a,m), \Tilde{\pi}(a,m,g) \rbrace_{a\in A, m\in M}$ for the gender $g\in\lbrace \text{male, female} \rbrace$. The quantity $a_{\max}:=121$ indicates the maximum, attainable age in the table.

\paragraph{Data preprocessing.} The categorical features 'gender' and 'smoker' are one-hot encoded. The payment style $m$ is treated as a numeric feature, motivated by the assumption of uniformly distributed, sub-annual death events. In line with common practice, see e.g. \cite{Wuethrich.2018}, we scale all features in contracts $c$ to a range of $[0,1]$ using min-max scaling, before we eventually apply an iterative stochastic gradient descent algorithm. For simplicity, we do not indicate the scaling explicitly in our notation, but assume from now on $c$ to be scaled. Further, as our objective is information decompression and not prediction, we do not employ a validation or test split on the data $\mathcal{D}$. However, in the upcoming Section \ref{section:model} we introduce an intrinsic and economic validation of the model performance.

%% file: Sections/Model.tex
\section{Neural network architecture} \label{section:model}
In the present use case, the state space $\mathcal{S}$ consists of two states, the policyholder being alive, encoded by $s_0:=0\in\mathcal{S}$, and them being dead, encoded by $1\in\mathcal{S}$. The initial state 'alive' is the only non-terminal state. Transition probabilities implicitly depend on the contract features, e.g. the age of the individual, and potentially the current iteration, which makes them inhomogeneous. By definition, the true probabilities satisfy $p_{10}^{(k)}(c):= 0$ and $p_{11}^{(k)}(c):= 1$ for any $k\in\mathbb{N}_0$ and $c\in\mathcal{C}$. Hence, we will we only need to model the first column of $\pi^{(k)}(c)=(p_{ij}^{(k)}(c))_{ij}$ explicitly. In the following, we present a suitable neural architecture, details on how to train it and our specific results of a hyperparameter tuning via 'hyperopt', see \cite{bergstra.2015} .

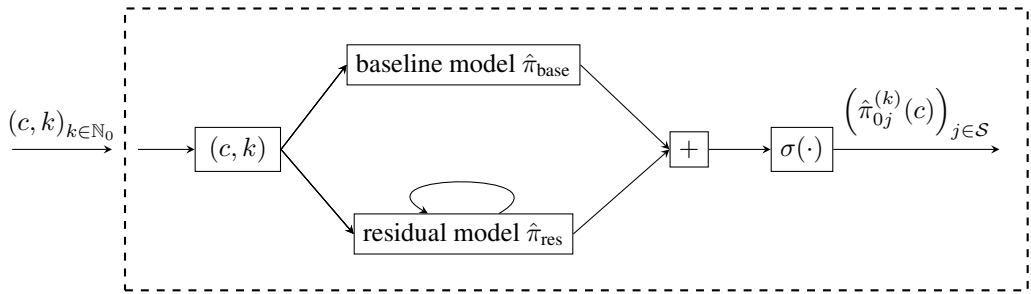
\begin{figure}[htb]
    \centering
    \input{Sections/model_architecture}
    \caption{Illustration of the overall architecture of $\hat{\pi}$, including sub-models $\hat{\pi}_{\text{base}}$ and $\hat{\pi}_{\text{res}}$ and a softmax activation function $\sigma(\cdot)$. The dashed box shows the computation for a fixed iteration $k\in\mathbb{N}_0$. The loop indicates the recurrent nature of $\hat{\pi}_{\text{res}}$ and the state $0\in\mathcal{S}$ is the single non-terminal state.}
    \label{fig:architecture:general}
\end{figure}

\paragraph{General architecture.}
We choose a recurrent network architecture with two sub-models $\hat{\pi}_{\text{base}}$ and $\hat{\pi}_{\text{res}}$, which in sum output a sequence of transition probabilities $(\hat{\pi}_{0j}^{(k)}(c))_{j=0,1}:=\sigma(\hat{\pi}_{\text{base}}(c,k)+\hat{\pi}_{\text{res}}(c,k)),~k\in\mathbb{N}_0$. The softmax activation function $\sigma(\cdot)$ ensures the output to be interpretable as true probabilities. \\
For the sub-model $\hat{\pi}_{\text{base}}$ we choose a comparably simple, feed-forward network that provides a meaningful baseline. Input features are restricted to a subset of $(c,k)$, namely the age at iteration $k$ and the payment style, but not e.g. the gender or the smoker status. We then train the sub-model $\sigma(\hat{\pi}_{\text{base}})$, subject to a softmax-activation $\sigma(\cdot)$, to fit homogeneous transition probabilities $\Tilde{\pi}\in\mathcal{F}$ from an appropriate mortality table. We select the German DAV 2008T table \cite{DAV:2008T}, which is commonly used for life insurances, and separately investigate both male and female transition probabilities as our baseline probabilities. This first step can be viewed as pre-training on a related task and will enable us to apply transfer learning, see e.g. \cite{Goodfellow.2016, Zhuang.2020}. \\
Keeping the baseline $\hat{\pi}_{\text{base}}$ fixed, the recurrent neural network $\hat{\pi}_{\text{res}}$ then utilizes more features of contract $c$ 
and fine tunes the joint model $\hat{\pi}$ for the semi-observable data set $\mathcal{D}=\lbrace c, \hat{y}(c) \rbrace_{c\in\mathcal{C}}$, recall Section \ref{section:data}. The recurrent structure of $\hat{\pi}_{\text{res}}$ allows for inhomogeneous transition probabilities $p_{ij}^{(k)}(c)$ to depend on information of the current path. Overall, we can think of the model $\hat{\pi}$ as a boosting machine, see e.g. \cite{Hastie.2017}. 
We illustrate the very general architecture in Figure \ref{fig:architecture:general}.  

\begin{remark}
\begin{itemize}
    \item[1.] This two-step approach is primarily motivated by the high dimensional setting with a rather small number of $N=10,000$ data points. The importance of the initialization, see e.g. \cite{Hanin.2018, Montufar.2020}, and the benefit of transfer learning, e.g. \cite{Zhuang.2020}, are well known. A poor initialization may lead to a failure in training and can in any case introduce an implicit bias during training with gradient descent. In preliminary experiments, we in fact observed the failure of training $\hat{\pi}_{\text{res}}$ in absence of $\hat{\pi}_{\text{base}}$. Further, recent literature \cite{Kiermayer.2021a, Gabrielli.2020, Horvath.2021} show the value of nested and modular architectures in terms of explainability and efficiency.
    \item[2.] If the dimension $\vert\mathcal{S}\vert$ was increased, for additional terminal states such as surrender, see e.g. \cite{Kiermayer.2021b}, we simply have to increase the width of the output layers of $\hat{\pi}_{\text{base}}$ and $\hat{\pi}_{\text{res}}$. For non-terminal states, such as disability, we have to create additional softmax-activated output layers, which each represents a column in the transition matrix $\hat{\pi}(c,k)$. The training can then be processed analogously in the style of the objective \eqref{eq:objective}.
    \item[4.] The type of recurrence in $\hat{\pi}_{\text{res}}$ effectively determines the limit of the path dependency in $(\hat{\pi}^{(k)}(c))_{k\in\mathbb{N}_0}$.  Details on different choices like a basic recurrent unit, GRU or LSTM can be found in \cite{Goodfellow.2016, cho.2014, chung.2014, hochreiter.1997}. 
    \item[4.] In the current setting with a single non-terminal state, where there exists only one path to each tuple of state and iteration, our architecture can also handle the state process $X(c)$ to be semi-Markovian or a Markov chain with memory. Detail on semi-Markov chains or Markov chains with memory can be found e.g. in \cite{Borovkov.2013, Klenke.2014, Wu.2017}. The generalization to more than one non-terminal state with an even more general assumption than (inhomogenous) Markovianity is an open research question.
\end{itemize}
\end{remark}

\paragraph{Baseline $\hat{\pi}_{\text{base}}$.} For the baseline model we transfer the mortality table DAV 2008T \cite{DAV:2008T} into a neural network. The data set $\mathcal{D}_{DAV}(g)= \lbrace (a,m), \Tilde{\pi}(a,m,g) \rbrace_{a\in A, m\in M}$, recall Section \ref{section:data}, contains the respective homogeneous transition probabilities $\Tilde{\pi}(a,m,g)$ for a fixed gender $g\in\lbrace \text{male}, \text{female}\rbrace$. The component $(a,m)$ in $\mathcal{D}_{DAV}(g)$ corresponds to the input of $\hat{\pi}_{base}$ and is limited to the payment style $m$ and the age $a$ at iteration $k$, which encodes information of the input tuple $(c,k)$ of the eventual overall model. Since we aim for a perfect fit of $\hat{\pi}_{\text{base}}$ on the full range of $\mathcal{D}_{DAV}$, overfitting is not an issue. Regardless, we still want to keep the complexity of $\hat{\pi}_{\text{base}}$ low since it will be a computational overhead once we fix it in step 2.\\

\begin{figure}[b!]
    \centering
    \input{Sections/sketch_architecture}
    \caption{General architecture of the sub-models $\hat{\pi}_{\text{base}}$ and $\hat{\pi}_{\text{res}}$ with depth $D$ and widths $n_i$ for layers $i=1,\ldots,D$, with $n_D=2$. In the ($D$-1)-th layer we implement either a dense feedfordward (FF) layer ($\hat{\pi}_{\text{base}}$) or a GRU layer ($\hat{\pi}_{\text{res}}$).}
    \label{fig:architecture:sub:model}
\end{figure}
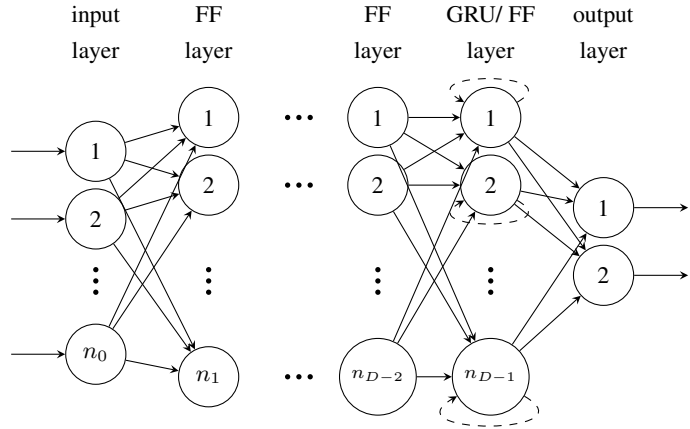
\begin{table}[b!]
    \centering
    \begin{tabular}{cccccccc|c}
    $D$ & $n_0$ & $n_1$ & $n_2$ & $n_3$ & $n_4$ & $B_{sz}$ & $l_{\text{rate}}$ & parameters \\
    \midrule
         4 & 2 & 40 & 40 & 20 & 2 & 32 & 0.001 & 2,622 \\
     \bottomrule
    \end{tabular}
    \caption{Hyperparameters of $\hat{\pi}_{base}$ and the 'adam'-optimizer, including the depth $D$, the width $n_i$ of the $i$-th layer, the learning rate $l_{rate}$ and the batchsize $B_{sz}$. We also state the number of parameters for the resulting architecture.}
    \label{tab:hparams:base}
\end{table}
We choose a $D$-layer dense, feedforward (FF) network, $D\in\mathbb{N}$, with ReLU activations, except for a final linear activation function. The width of the $i$-th layer is denoted by $n_i$. We display the architecture of $\hat{\pi}_{\text{base}}$ in Figure \ref{fig:architecture:sub:model}, where in the ($D$-1)-th layer the FF layer option applies. Given the softmax activation function $\sigma(\cdot)$, we then train the model $\sigma(\hat{\pi}_{\text{base}})$ with a Kullback-Leibler divergence loss function see e.g. \cite{Goodfellow.2016}. The stochastic gradient descent is performed by the 'adam' optimizer, see \cite{adam}. After a manual search we find appropriate hyperparameters, which are summarized in Table \ref{tab:hparams:base}. The corresponding model $\hat{\pi}_{base}$ has 2,622 trainable parameters and provides transition probabilities that are reasonably\footnote{A visualization can be found in the Appendix, 
Figure \ref{fig:baseline:fit}.} close to those in the DAV 2008T table, across the full range of ages and payment styles in the data set $\mathcal{D}_{DAV}(g)$, $g\in\lbrace \text{male}, \text{female} \rbrace$. Recall that the estimate $\hat{\pi}_{\text{base}}$ of $\Tilde{\pi}$ is simply an educated, initial guess for the optimal value ${\pi}^{\star}$, see \eqref{eq:objective}, that can be corrected via $\hat{\pi}_{res}$.

\paragraph{Residual $\hat{\pi}_{\text{res}}$.} In the next step, we create the model $\hat{\pi}_{\text{res}}$ to boost and, thereby, correct any ill-fitted baseline $\hat{\pi}_{\text{base}}$. The semi-observable data set $\mathcal{D}$ contains contracts $c\in\mathcal{C}$ and non-observable, discounted cash flows $\hat{y}(c)$, recall Section \ref{section:data}. Motivated by the goal of explainability, we restrict the input to $\hat{\pi}_{\text{res}}$ to $n_0=4$ features, namely the age at iteration $k$, the payment style $m$, the gender $g$ and the smoker status. Also, from a legal perspective the transition probabilities $\hat{\pi}^{(k)}(c)$ of contract $c$ at iteration $k$ should not be allowed to depend on e.g. the premium or the sum insured. \\
For the architecture of $\hat{\pi}_{\text{res}}$, we choose a network of depth $D$, where all layers previous to the $(D$-1)-th layer are dense, feedforward layers with a ReLU activation. The $(D$-1)-th layer contains gated-recurrent units (GRU) with a default hyperbolic tangens ($\tanh$) activation, see \cite{cho.2014}, followed by a linear, dense layer with $n_D:=2$ output units. The architecture of $\hat{\pi}_{\text{res}}$ is displayed in Figure \ref{fig:architecture:sub:model}. Based on experimental results\footnote{During an automated tuning with hyperopt \cite{bergstra.2015}, multiple GRU layers showed no improved performance and the coefficient of the $L_2$-regularization was tuned down to a small value. Dropping the regularization all together improved the performance.}, we abstain from using more than one GRU layer or a layer-wise $L_2$-regularization. The economic function $\psi(\cdot)$ already provides a practical restriction, that prevents even highly overparametrized neural network from simply memorizing target values. \\
To implement the boosting machine, we form the joint model $(\hat{\pi}_{0j}^{(k)}(c))_{j=0,1}:=\sigma(\hat{\pi}_{\text{base}}(c,k)+\hat{\pi}_{\text{res}}(c,k)),~k=1,\ldots,T$ for the maximum sequence length $T=576$, which corresponds to a monthly payment style ($m=12)$ over the maximum duration of $n=48$ years. At this point, we fix the baseline $\hat{\pi}_{base}$, calibrated either for male or female transition probabilities in the DAV 2008T table. In line with the objective \eqref{eq:objective} and common practice , see e.g. \cite{Goodfellow.2016}, we then minimize the empirical risk
\begin{align*}
    R_{emp}(\hat{\pi}) := \frac{1}{\vert \mathcal{D}\vert} \sum_{c,\hat{y}(c)\in\mathcal{D}} \ell\circ\psi(\hat{\pi},c,\hat{y}(c)).
\end{align*}
In all following experiments, we set $\ell:\mathbb{R}\rightarrow\mathbb{R}_{+}, z\mapsto \vert z \vert$ as the absolute\footnote{One may also consider a more general, symmetric function $\ell(z):=\vert z \vert^{p}$, $p>0$. However, values $p>1$ raise practical concerns during gradient based optimization due to the unlimited scale of discounted cash flows $\hat{y}(c)$.} value function. The minimization is performed via stochastic gradient descent and the algorithm 'adam', see \cite{adam}. In order to estimate optimal hyperparameters of the architecture of $\hat{\pi}_{res}$ and the training algorithm we combine a manual and an automated search, the later based on the python package 'hyperopt' and its Bayesian 'tpe.suggest' search algorithm, see \cite{bergstra.2015}. \\ 
Our experiments show that tuning the depth $D$, learning rate $l_{rate}$ and the batch size $B_{sz}$ outweigh the effect of other parameters such as the widths $n_i$. Originally, values $n_i\in\lbrace 20,30,\ldots, 60 \rbrace$, $i=2,\ldots, D$-$1$, were explored by 'hyperopt'. However, given the little benefit and high computational cost of tuning $n_i$, we fix sufficiently wide, hidden layers with $n_i:=50$ and explore depths $D\in\lbrace 4,5,6,7\rbrace$, where the final layer with its $n_D:=2$ output neurons is fixed. Further, all combinations of learning rates $l_{rate}\in \lbrace 0.001, 0.005, 0.01 \rbrace$ and batch sizes $B_{sz}\in\lbrace 32,64, 128 \rbrace$ per GPU\footnote{The training of neural models is distributed across eight Quadro RTX 8000.} are investigated. To promote convergence, we include a schedule that reduces the learning rate by $10\%$ every $15$ epochs after a warm-up period of $50$ epochs and early stopping with a patience of $50$ epochs. 
The results of the hyperparameter search are displayed in Table \ref{tab:hparams:res} and show a very similar setting for both genders, aside from a lower learning rate for the female baseline.
\begin{table}[htb]
    \centering
    \begin{tabular}{c|cccccccccc|c}
       baseline & $D$ & $n_0$ & $n_1$ & $n_2$ & $n_3$ & $n_4$ & $n_5$ & $n_6$ & $B_{sz}$ & $l_{rate}$ & parameters \\
       \midrule
       male & 6 & 4 & 50 & 50 & 50 & 50 & 50 & 2 & 32 & 0.005 & 23,302\\
       female & 6 & 4 & 50 & 50 & 50 & 50 & 50 & 2 & 32 & 0.001 & 23,302 \\
       \bottomrule
    \end{tabular}
    \caption{Hyperparameters of $\hat{\pi}_{res}$ and resulting number of trainable parameters. The baseline column indicates the gender $g\in\lbrace \text{male},\text{female} \rbrace$ which the fixed baseline $\hat{\pi}_{base}$ was calibrated on.}
    \label{tab:hparams:res}
\end{table}

\begin{remark}
    \begin{itemize}
        \item[1.] While traditionally overparametrization and the curse of dimensionality are common concerns when fitting a model, see e.g. \cite{Hastie.2017, Bishop.2016}, current literature, such as \cite{Nakkiran.2020, AllenZhu.2019, Du.2019, Arora.2018}, shows (highly) overparametrized neural models to perform well in terms of learning and generalization. Our model is moderately overparametrized in the sense that the number of parameters exceeds the number of data points $N=10,000$ by a factor of $2.5$. Although our objective is explainable, local information retrieval and not a generalization outside of the populated feature space, we still find overparametrization to be beneficial. For depths $D<6$, the economic model validation differs from our results with $D=6$, which will be presented in Section \ref{section:num_experiments}, primarily in the magnitude and quantity of outliers.  Therefore, our experiments indicate overparametrization and depth to promote more robust estimates $\hat{\pi}(\cdot)$. Depths $D>6$ showed no futher improvements.
        \item[2.] Effectively, we aim to estimate about $37,632 =42\cdot 4\cdot 2\cdot 2\cdot 14\cdot 4$ values ${\pi}_{01}^{(k)}(c)$ from $N=10,000$ data points. This number corresponds to the number of possible values for the initial age $a_0$, the payment style $m$, the gender $g$ and the smoker status recorded in $c$, as well as mean median duration $n$ and median payment style $m$, which in combination determine the number of iterations $k=0,1,\ldots,nm$-$1$ for a single policy $c$. Assuming homogeneity would reduce the number to $672=42\cdot 4\cdot 2\cdot 2.$
        \item[3.] For the training of the joint model $\hat{\pi}$, we encode the pair $(c,k)$ of contract $c$ at iteration $k$ by adjusting the current age by $a_0+km$, where $a_0$ indicates the initial age and $m$ the payment style, e.g. monthly ($m=12$). Then all input data are stored as a sequence of length $T=576$, using zero-padding if applicable.
        \item[4.] To speed up training, we apply masking to ignore zero-padded time steps in the loss function, although these time steps $k$ have discounted cash-flows $\hat{y}^{(k)}(c)=0 \in\mathbb{R}^{2\times2}$ anyway and, thus, do not contribute to the empirical risk $R_{emp}$.
        \item[5.]During the training of $\hat{\pi}$, we observed exploding gradients for large learning rates, presumably due to the unbounded nature of cash flows $y(c)$ in combination with a recurrent structure. To mitigate this issue, we apply gradient clipping with a value of $\eta=100$, see e.g. \cite{Goodfellow.2016, Quian.2021}.
    \end{itemize}
\end{remark}

\paragraph{Intrinsic model validation.} Recall that our main objective is information retrieval on the given, small data set, which is why we do not utilize a validation or test set. However, once we have calibrated our model and obtained estimates $\hat{\pi}$, a very natural approach is to backtest the premium values $P$, recorded in the data set $\mathcal{D}$. Given transition probabilities $\hat{\pi}$, we can compute estimates $\hat{P}$ that result in APV-consistent contracts $c$. Analogously to Section \ref{section:data}, for a clearer display we use separate letters instead of indexing to indicate features in $c$.\\

Let us assume an APV-consistent contract $c$ which satisfies our assumptions \eqref{eq:cf:active} - \eqref{eq:cf:discounted} for its cash flows. Further, observe that the cash flows can be separated in summands related to the premium $P\in\mathbb{R}_{\geq 0}$ and the sum insured $S\in\mathbb{R}_{\geq 0}$, respectively. One may think of $y(c)$ as a function $g(P,S):=y(c)$, for which holds $g(P,S)=g(P,0)+g(0,S)$. By \eqref{eq:psi}, the function $\psi$ is linear with respect to its the discounted cash flows $y(c)$. Therefore, by the APV-consistency of $c$ the true quantity $\pi^{\star}$ satisfies
\begin{align*}
    0 :&= \psi(\pi^{\star}, c,y(c)) \\
        &= \psi(\pi^{\star}, c,y(c))\bigg\rvert_{P=0} + \psi(\pi^{\star}, c,y(c))\bigg\rvert_{S=0},
\end{align*}
where the condition '$\rvert_{S=0}$', resp. '$\rvert_{P=0}$', corresponds to dropping the features in $c$ and cash flows in $y(c)$ related to the sum insured and the premium, respectively.  Further, explicitly indicating the premium related quantities in \eqref{eq:cf:active} - \eqref{eq:cf:discounted} in the function $\psi$ yields
\begin{align*}
    \psi(\pi^{\star}, c,y(c))\bigg\rvert_{S=0} &:=  - \frac{P}{m}t\alpha + \frac{P}{m}(1-\beta)\sum_{k=0}^{tm-1}v^{\frac{k}{m}} M^{(0,k-1)}_{00}(c)\hat{\pi}^{(k-1)}_{00}(c).
\end{align*}
Hence, the APV-consistently calibrated premium $P$ for contract $c$ is given by
\begin{align} \label{eq:premium:backtest}
    P := \psi(\pi^{\star}, c,y(c))\bigg\rvert_{P=0} \left(\frac{t\alpha}{m}-(1-\beta)\sum_{k=0}^{tm-1}\frac{1}{m}v^{\frac{k}{m}} M^{(0,k-1)}_{00}(c)\hat{\pi}^{(k-1)}_{00}(c)\right)^{-1}.
\end{align}
As desired, the right hand side of \eqref{eq:premium:backtest} does not depend on $P$. To evaluate the calibrated model $\hat{\pi}$, we can therefore backtest the actual, recorded premium value $P$ with an estimate $\hat{P}$ by simply inserting our corresponding estimates in \eqref{eq:premium:backtest}. Note that the estimate $\psi(\hat{\pi}, c,\hat{y}(c))\bigg\rvert_{P=0}$ can be computed efficiently with the calibrated architecture $\hat{\pi}$ by simply dropping dependencies on $P$ from both $c$ and $\hat{y}(c)$, i.e.  effectively setting the value to zero, resp. its equivalent, scaled value. For actuaries it might be helpful to note that the sum in \eqref{eq:premium:backtest} corresponds to an annuity value, commonly denoted by $\text{ä}_{x:\angl{n-t}}^{(m)}$, see e.g. \cite{Fuehrer:2010, dickson:2009}.

%% file: Sections/model_architecture.tex
\tikzset{%
  every neuron/.style={
    circle,
    draw,
    minimum size=0.5cm
  },
  neuron missing/.style={
    draw=none, 
    scale=2,
    text height=0.333cm,
    execute at begin node=\color{black}$\vdots$
  },
}

    \begin{tikzpicture}[x=1.5cm, y=1.5cm, >=stealth]
        \tikzstyle{rectangle_style}=[rectangle, draw, minimum size = 0.4cm]
        \tikzstyle{bigbox} = [draw, thick, rounded corners, rectangle]
        
        \node[draw, circle, color = white] (input) at (0,0) {};
        
        \draw [<-] (input) -- ++(-1,0) 
            node [above, midway] {$\left(c, k\right)_{k\in\mathbb{N}_0}$};
            
        \node[draw, rectangle_style, text width=0.9cm,minimum height=0.5cm, minimum width=0.5, align = center] (input-layer) at 
     (1,0) {$\left(c, k\right)$};
        \draw [->] (input) to (input-layer);

        \node[draw, rectangle_style, minimum width=2] (baseline) at 
     (3,0.75) {baseline model $\hat{\pi}_{\text{base}}$};
        \node[draw, rectangle_style, minimum width=2] (residual) at 
     (3,-0.75) {residual model $\hat{\pi}_{\text{res}}$};
        \draw [->] (residual) to [out=30,in=150,looseness=3] (residual);
        
         \foreach \m/\l [count=\y] in {1,2,n}
                \draw [->] (input-layer) to [out=0,in=180, looseness=0] (baseline);
        \foreach \m/\l [count=\y] in {1,2,n}
            \draw [->] (input-layer) to [out=0,in=180, looseness=0] (residual);
        
        \draw node at (5, 0.) [rectangle_style] (add) {$+$};
        \draw node at (6, 0.) [rectangle_style] (actv) {$\sigma(\cdot)$};
        
        \draw [->] (baseline) to [out=0,in=180, looseness=0] (add);
        \draw [->] (residual) to [out=0,in=180, looseness=0] (add);
        \draw [->] (add) to (actv);
        
        \draw [->] (actv) -- ++(1.75,0) node [above, midway] {$\left(\hat{\pi}_{0j}^{(k)}(c)\right)_{j\in\mathcal{S}}$};

        \draw[black,thick, dashed] (0, 1.25)  rectangle (8,-1.25);
        

    \end{tikzpicture}

%% file: Sections/sketch_architecture.tex
\tikzset{%
  every neuron/.style={
    circle,
    draw,
    minimum size=0.8cm
  },
  neuron missing/.style={
    draw=none, 
    scale=2,
    text height=0.333cm,
    execute at begin node=\color{black}$\vdots$
  },
}

    \begin{tikzpicture}[x=1.5cm, y=1.5cm, >=stealth]
        \tikzstyle{rectangle_style}=[rectangle, draw, minimum size = 0.4cm]
        
        \foreach \m/\l [count=\y] in {1,2}
          \node [every neuron/.try, neuron \m/.try] (input-\m) at (1,1.9-\m*0.6) {\footnotesize{\m}};
        \foreach \m/\l [count=\y] in {n}
          \node [every neuron/.try, neuron \m/.try] (input-\m) at (1,-0.5) {\footnotesize{$n_0$}};
         \foreach \x in {1,...,3}
            \fill (1, 0.35 -\x*0.1) circle (1pt); 
            
        \draw [<-] (input-1) -- ++(-0.75,0);
        \draw [<-] (input-2) -- ++(-0.75,0);
        \draw [<-] (input-n) -- ++(-0.75,0);
            
        \node [align=center, above] at (1, 2) {\footnotesize{input}\\ \footnotesize{layer}};

        \foreach \m/\l [count=\y] in {1,2}
          \node [every neuron/.try, neuron \m/.try] (dense0-\m) at (2,2.2-\m*0.6) {\footnotesize{\m}};
        \foreach \m/\l [count=\y] in {n}
          \node [every neuron/.try, neuron \m/.try] (dense0-\m) at (2,-0.7) {\footnotesize{$n_1$}};
        \foreach \x in {1,...,3}
            \fill (2, 0.35 -\x*0.1) circle (1pt);
        \node [align=center, above] at (2, 2) {\footnotesize{FF}\\ \footnotesize{layer}};
        
        \foreach \m/\l [count=\y] in {1,2}
          \node [every neuron/.try, neuron \m/.try] (dense1-\m) at (3.5,2.2-\m*0.6) {\footnotesize{\m}};
        \foreach \m/\l [count=\y] in {n}
          \node [every neuron/.try, neuron \m/.try] (dense1-\m) at (3.5,-0.7) {\scriptsize{$n_{D-2}$}};
        \foreach \x in {1,...,3}
            \fill (3.5, 0.35 -\x*0.1) circle (1pt);
        \node [align=center, above] at (3.5, 2) {\footnotesize{FF}\\ \footnotesize{layer}}; 
        \foreach \x in {1,...,3}
            \fill (2.6+\x*0.1, 1.6) circle (1pt);
        \foreach \x in {1,...,3}
            \fill (2.6+\x*0.1, 1) circle (1pt);
        \foreach \x in {1,...,3}
            \fill (2.6+\x*0.1, -0.7) circle (1pt);
        
        \foreach \m/\l [count=\y] in {1,2}
          \node [every neuron/.try, neuron \m/.try] (gru-\m) at (4.5,2.2-\m*0.6) {\footnotesize{\m}};
        \foreach \m/\l [count=\y] in {n}
          \node [every neuron/.try, neuron \m/.try] (gru-\m) at (4.5,-0.7) {\scriptsize{$n_{D-1}$}};
        \foreach \x in {1,...,3}
            \fill (4.5, 0.35 -\x*0.1) circle (1pt);
        \node [align=center, above] at (4.5, 2) {\footnotesize{GRU/ FF}\\ \footnotesize{layer}}; 
        \draw [->, dashed] (gru-1) to [out=30,in=150,looseness=3] (gru-1);
        \draw [->, dashed] (gru-2) to [out=330,in=210,looseness=3] (gru-2);
        \draw [->, dashed] (gru-n) to [out=330,in=210,looseness=3] (gru-n);

        \foreach \m/\l [count=\y] in {1,2,n}
            \foreach \k in {1,2,n}
                \draw [->] (input-\m) -- (dense0-\k);
        \foreach \m/\l [count=\y] in {1,2,n}
            \foreach \k in {1,2,n}
                \draw [->] (dense1-\m) -- (gru-\k);

        \foreach \m/\l [count=\y] in {1,2}
          \node [every neuron/.try, neuron \m/.try] (dense3-\m) at (5.5,1.4-\m*0.6) {\footnotesize{\m}};
        \node [align=center, above] at (5.5, 2) {\footnotesize{output}\\ \footnotesize{layer}};


        \foreach \m/\l [count=\y] in {1,2,n}
            \foreach \k in {1,2}
                \draw [->] (gru-\m) -- (dense3-\k);

        
                
                
            
                
        \draw [->] (dense3-1) -- ++(0.75,0) 
            node [above, midway] {};
        \draw [->] (dense3-2) -- ++(0.75,0) 
            node [above, midway] {};
            

    \end{tikzpicture}

%% file: Sections/Numerical_Results.tex
\section{Numerical Results} \label{section:num_experiments}
Next, we provide the results for our methodology applied to the data set of $N=10,000$ German term life insurance contracts, provided by msg life central europe gmbh. First, we present a qualitative description of the obtained transition probabilities, which might be interpreted as a diagnostic tool for risk types of policyholders. However, this qualitative presentation relies on the correctness of our approach and allows for little reasoning yet to explain the results. Second, we quantitatively verify the extracted Markov chain by backtesting observed premium values, recall our intrinsic model validation in Section \ref{section:model}. Both steps will be performed for a baseline of male and female transition probabilities from the DAV 2008T table \cite{DAV:2008T} separately and, thereby, testing the stability of our approach with respect to its initialization.

\begin{figure}[bp]
        \centering
        \begin{subfigure}{.49\textwidth}
            \centering
            \includegraphics[width=\textwidth]{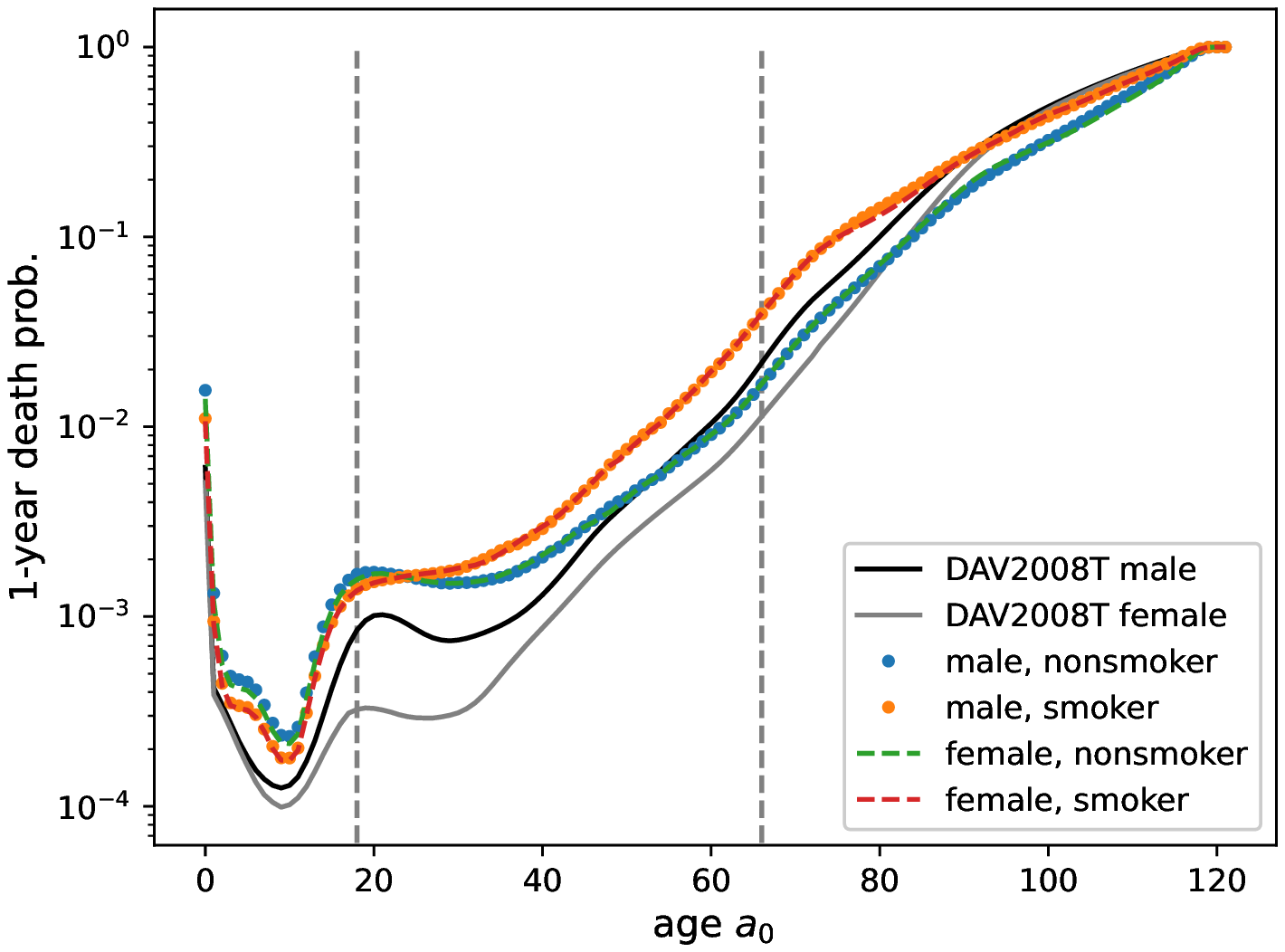}
            \subcaption{$\hat{\pi}_{\text{base}}$ calibrated for DAV2008T male.}
            \label{fig:implied:mortalitiy:male}
        \end{subfigure}
        \begin{subfigure}{.49\textwidth}
            \centering
            \includegraphics[width=\textwidth]{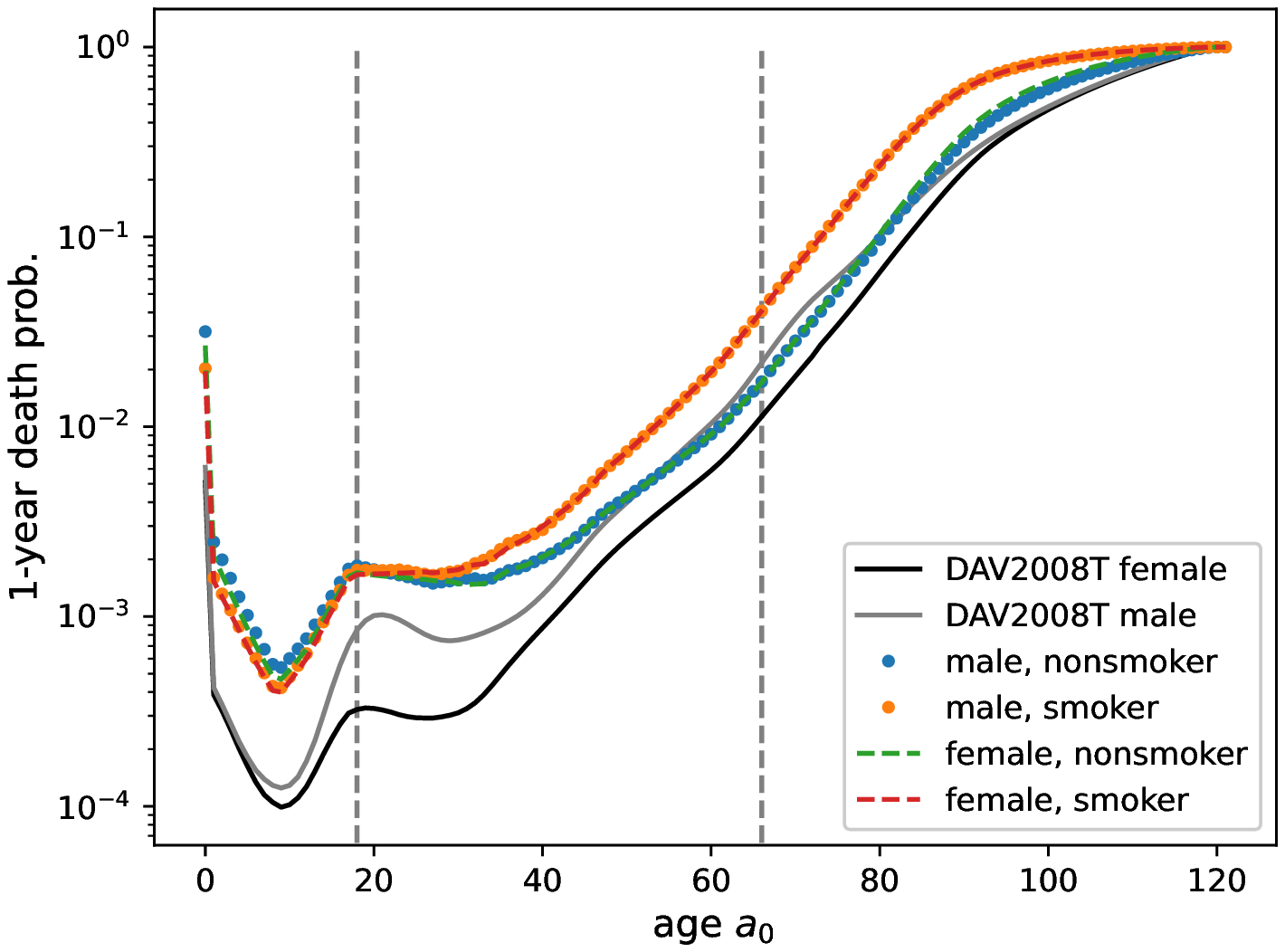}
            \subcaption{$\hat{\pi}_{\text{base}}$ calibrated for DAV2008T female.}
            \label{fig:implied:mortalitiy:female}
        \end{subfigure}
    \caption{Mortality curves $\hat{\pi}^{(0)}_{01}(c)$ for different (initial) ages and combinations of gender and smoker status.}
    \label{fig:implied:mortality}
\end{figure}

\paragraph{Implied mortality.} We start with a descriptive analysis by comparing the mortality curve of the DAV2008T table to the mortality $\hat{\pi}^{(k)}(c)$ implied by our calibrated model $\hat{\pi}$ for contract $c\in\mathcal{C}$ at times $k=0$ and for an annual payment style $m=1$. This will keep the illustration comprehensible. In subsequent steps we will check for all payments styles $m$ and for the effect of $k\in\mathbb{N}_0$, which can indicate an inhomogeneity of the Markov chain. \\
In Figure \ref{fig:implied:mortality}, we see the mortality curves recorded of the DAV2008T table for male as well as female individuals, and the $1$-year death probabilities $\hat{\pi}^{(0)}_{01}(c)$ of two fully trained models $\hat{\pi}$ for different initial ages $a_0$ recorded in contracts $c\in\mathcal{C}$. The two subplots differ by which gender was used for the training of the baseline $\hat{\pi}_{\text{base}}$. The reference DAV table of the respective gender is highlighted in black. The dotted, vertical lines indicate the maximum range of ages from $18$ to $66$ that policyholders of $c\in\mathcal{C}$ show throughout the full duration of their contract. \\
\begin{figure}[b!]
    \centering
    \begin{subfigure}{\textwidth}
        \centering
        \includegraphics[width=0.7\textwidth]{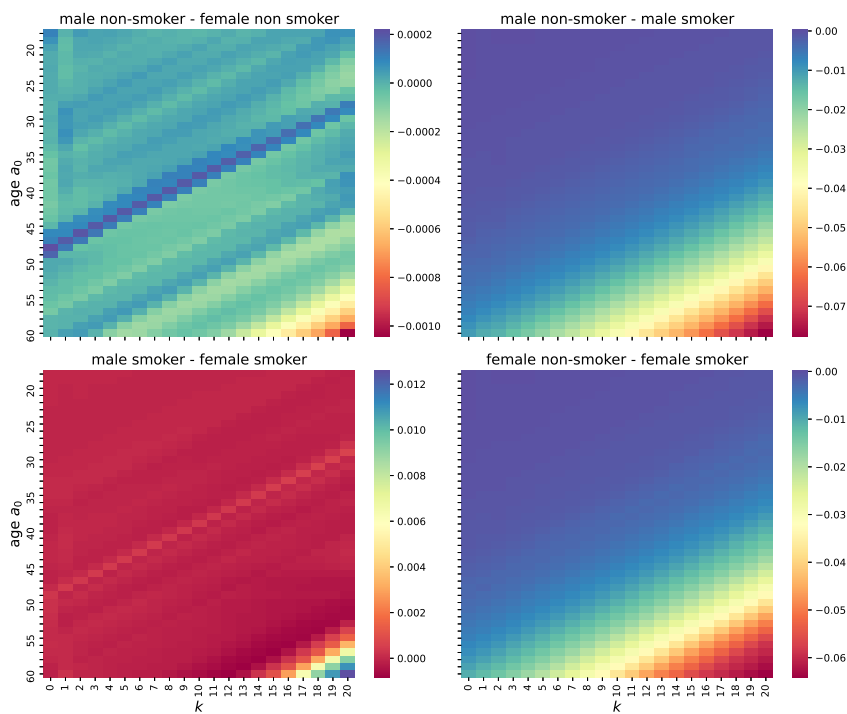}
        \subcaption{With $\hat{\pi}_{\text{base}}$ calibrated for DAV2008T male and payment style $m=1$. }
        \label{fig:mortality:heatmap:male:zoom}
    \end{subfigure}
    \caption{Difference of mortality probabilities $\hat{\pi}_{01}(c)$ with different gender and smoker status in contract $c$.}
\end{figure}
\begin{figure}[tbh] \ContinuedFloat
    \centering
    \begin{subfigure}{\textwidth}
    \centering
        \includegraphics[width=0.8\textwidth]{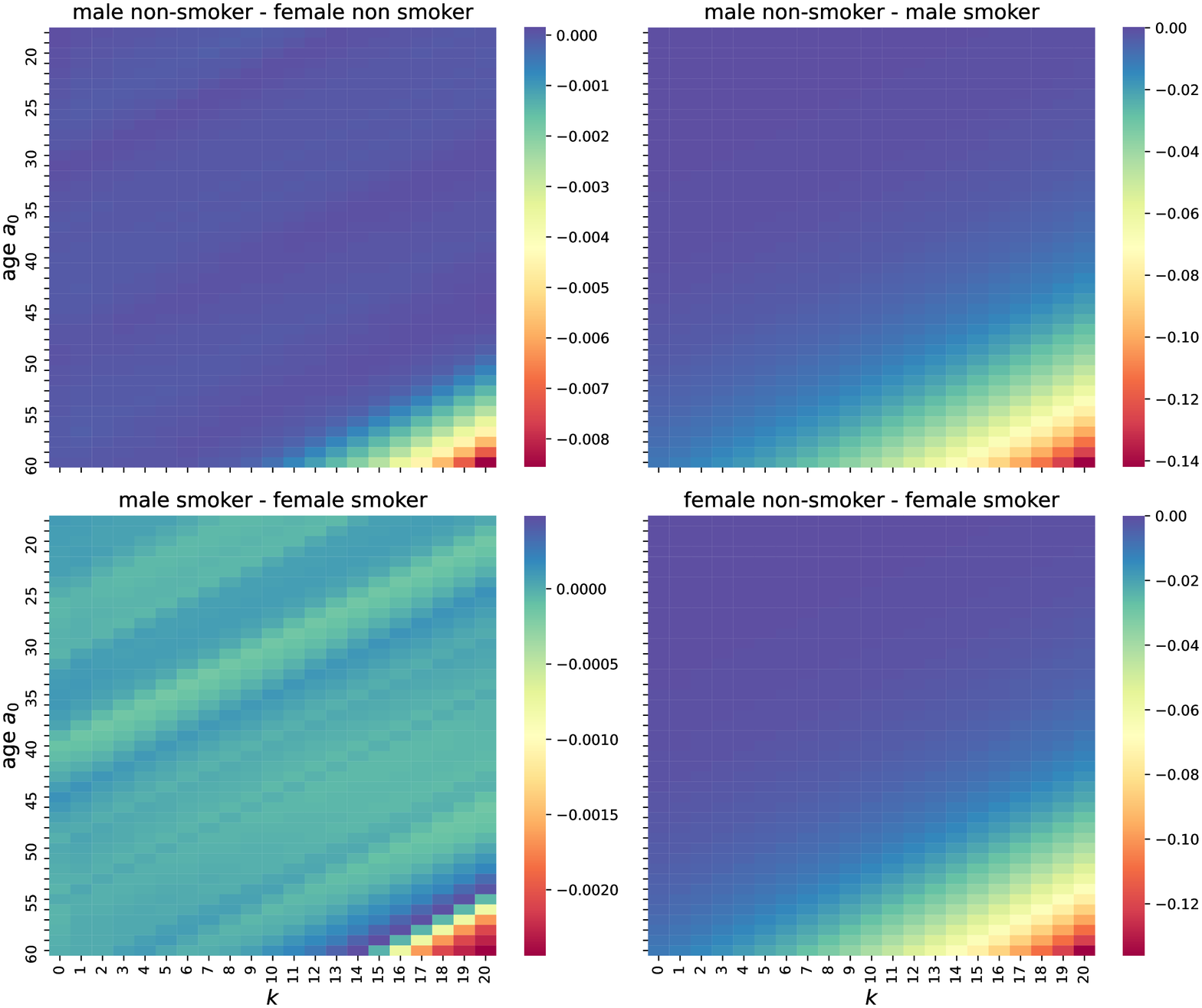}
        \subcaption{With $\hat{\pi}_{\text{base}}$ calibrated for DAV2008T male and payment style $m=1$. }
        \label{fig:mortality:heatmap:female:zoom}
    \end{subfigure}
    \captionsetup{list=off,format=cont}
    \caption{Difference of mortality probabilities $\hat{\pi}_{01}(c)$ with different gender and smoker status in contract $c$.}
    \label{fig:mortality:heatmap:zoom}
\end{figure}
We would like to point out some main observations in Figure \ref{fig:implied:mortality}. First, the calibrated probabilities $\hat{\pi}^{(0)}(c)$ differ mainly by the smoker status and are hardly affected by the gender of the policyholder. This holds true regardless of which gender the baseline $\hat{\pi}_{\text{base}}$ was trained on. Since generally male mortality rates are higher, see e.g. \cite{DAV:2008T}, this is surprising and can only be explained by an underlying unisex tariff. In hindsight, we were able to confirm with msg central life europe gmbh that the data indeed stems from a unisex tariff. A model based evaluation and confirmation will be presented in the next paragraph. Second, quantities $\hat{\pi}^{(0)}(c)$ generally exceed the common DAV2008T baseline on the populated range of ages from $18$ to $66$. Starting at age $50$ the non-smoker rates approximately coincide with the male DAV2008T table. Further, for ages that are not covered in our data $\mathcal{D}$, i.e. outside of the range $18$ to $66$, we observe significant disparities between initializations. For high ages the death probability of smokers appears unreasonably high compared to non-smokers, especially when $\hat\pi$ was calibrated for a male baseline as in Figure \ref{fig:implied:mortalitiy:male}, where the death probability $\hat{\pi}^{(0)}_{01}(c)$ at the age of $80$ is doubled by being a smoker. While it is important to be aware of a limited generalization, this behaviour is expected and does not conflict with our objective of local information retrieval. At the same time, we clearly see how the initialization affects the general shape of the mortality curves. In particular,  for lower ages outside of the populated area of the feature space the calibrated curves are, on the logarithmic scale, an approximately shifted version of the baseline despite of the absence of data for that subset. 
This final observation indicates the general benefit of transfer learning and is in line with theoretical findings\footnote{To be precise, findings in \cite{Montufar.2020} are shown formally for (non-stochastic) gradient descent and shallow ReLU networks and do not include recurrent GRU layers. Our observations suggest that this might also hold more generally.} in \cite{Montufar.2020}, where gradient descent is biased towards keeping smoothness properties of the initialization. \\

Next, we investigate time steps $k>0$ of the underlying Markov chain. For comprehensiveness, our illustrations are restricted to annual payments $m=1$ and the range of initial ages $a_0=18,19,\ldots,60$ populated by our data $\mathcal{D}$. Motivated by the observation in Figure \ref{fig:implied:mortality}, we further choose to investigate the difference $\hat{\pi}^{(k)}_{01}(c)-\hat{\pi}^{(k)}_{01}(c^{\prime})$, where contracts $c,c^{\prime}$ only differ in the quantitative features gender and smoker-status. In Figure \ref{fig:mortality:heatmap:zoom}, these differences are displayed for a male and a female baseline model $\hat{\pi}_{\text{base}}$. Each row in the heatmap corresponds to a different value $a_0$ of contracts $c, c^{\prime}$. Note that each heatmap has an individual colorbar due to differing scales. The title indicate the quantitative features by which $c$ and $c^{\prime}$ differ. \\
In Figure \ref{fig:mortality:heatmap:zoom}, we observe that all heatmaps are approximately constant across diagonals. This strongly suggests that the calibrated transition probabilities of the underlying Markov chain are homogeneous, if the current age $a_0+km$ is included in the state space. We can easily confirm this by checking not the difference, but individual mortality probabilities. The corresponding heatmaps for $\hat{\pi}(c)$ mirror Figure $\ref{fig:mortality:heatmap:zoom}$ and confirm the approximate homogeneity for all combinations of gender and smoker-status. An examplary illustrating for the female baseline model is provided in the Appendix, Figure \ref{fig:mortality:homogeneity:female}. 
Further, the heatmaps also show a unisex tariff with absolute discrepancies of less than $10^{-3}$ on the populated range of current ages $a_0+km\in\lbrace 18,\ldots,66\rbrace$, see the first column in Figures \ref{fig:mortality:heatmap:male:zoom} and \ref{fig:mortality:heatmap:female:zoom}, as well as a increased mortality probability for smokers, see the non-positive colorbar in the second column in Figures \ref{fig:mortality:heatmap:male:zoom} and \ref{fig:mortality:heatmap:female:zoom}. 
Last, we very clearly observe the boundary of the populated feature space and the limitation of our information retrieval algorithm. In the data, the maximum initial age $a_0$ equals $60$ and the maximum value of $a_0+km$ equals $66$. Starting at values in the range of $66$ to $74$, the heatmaps show increasing differences between the combinations of levels of quantitative features. The magnitudes of these differences are both unrealistic and impractical, but are to be expected and stem from the local\footnote{In practice, one may think about regularizations and model designs that restrict the deviation of $\hat{\pi}^{(k)}_{01}(c)$ from a reference value, such as a DAV table, to a fixed, maximum value. While this, by design, would mitigate deviations outside of the populated feature space, without data to reliably and explainably validate probabilities $\hat{\pi}^{(k)}_{01}(c)$ there will be little practical benefit.} nature of our approach. In the Appendix in Figure \ref{fig:mortality:heatmap:full}, we reproduce the information of Figure \ref{fig:mortality:heatmap:zoom} for a wider range of initial ages $a_0$ and again confirm that outside of the populated regions of the feature space, calibrated transition probabilities present an impractical 
extrapolation.\\
Overall, Figure \ref{fig:mortality:heatmap:female:zoom} confirms our observations in Figure \ref{fig:implied:mortality} from a more general perspective. Due to the approximate homogeneity, single step transition probabilities $\hat{\pi}^{(k)}_{01}$ at ages $a_0$ and iteration $k=0$ in Figure \ref{fig:implied:mortality} show a representative mortality curve for iterations $k\in\mathbb{N}_0$, when we include the current age $a_0+km$ in the state space of the underlying Markov process.

\begin{figure}[bph]
        \centering
        \begin{subfigure}{.45\textwidth}
            \centering
            \includegraphics[width=\textwidth]{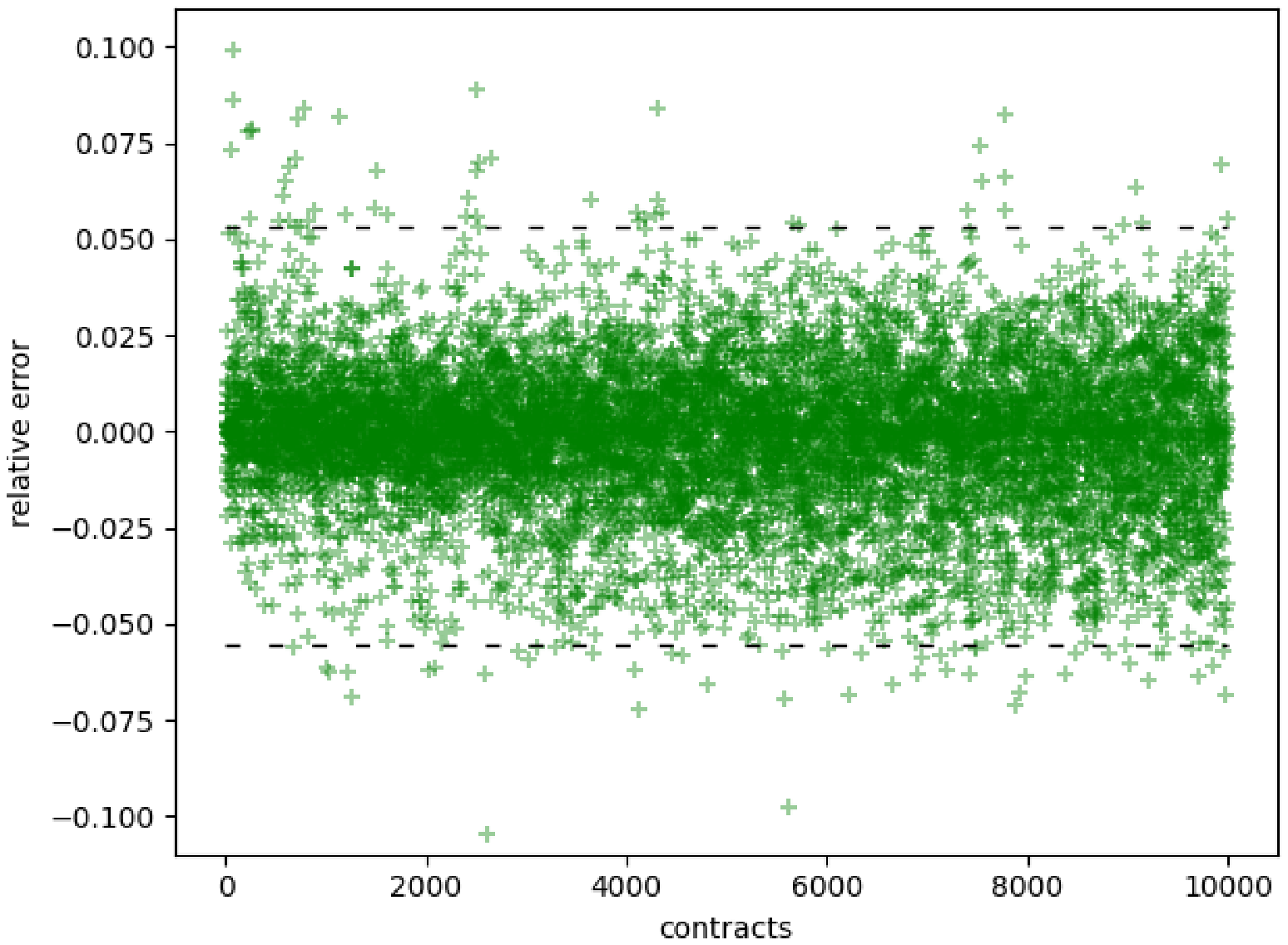}
            \subcaption{With $\hat{\pi}_{\text{base}}$ calibrated for DAV2008T male.}
            \label{fig:errors:relative:male}
        \end{subfigure}
        \begin{subfigure}{.49\textwidth}
            \centering
            \includegraphics[width=\textwidth]{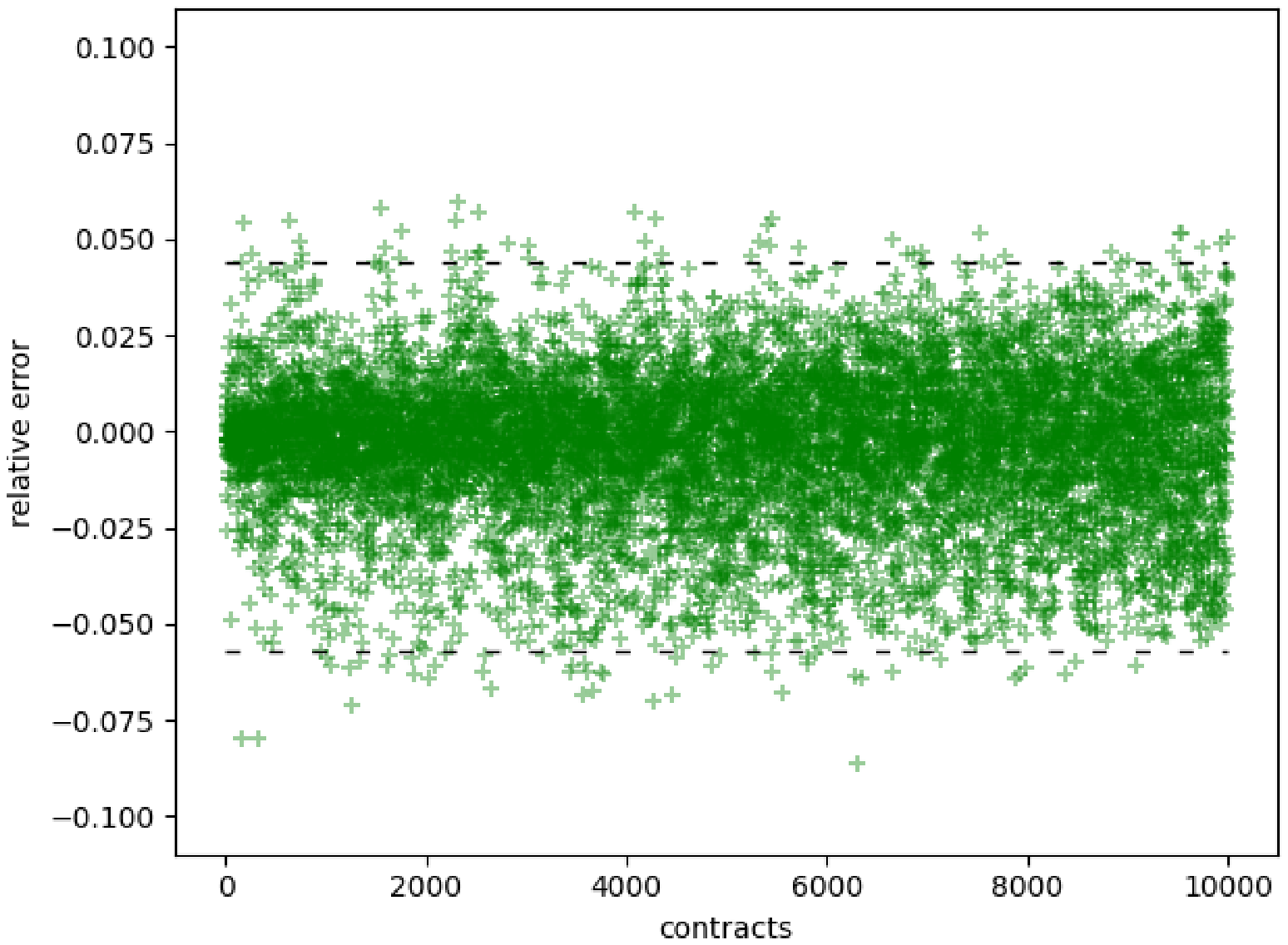}
            \subcaption{With $\hat{\pi}_{\text{base}}$ calibrated for DAV2008T female.}
            \label{fig:errors:relative:female}
        \end{subfigure}
    \caption{Relative errors $e_{\text{rel}}(c)$ for $c\in\mathcal{C}$. Horizontal lines indicate $\alpha$-quantiles $q_\alpha$ for $\alpha\in\lbrace 0.05, 0.995\rbrace$.}
    \label{fig:errors:relative}
\end{figure}

\paragraph{Economic validation.}
So far we have provided a qualitative analysis limited to a payment style $m=1$, for the sake of comprehensible illustrations. Next, we perform a quantitative intrinsic model validation of $\hat{\pi}$, recall Section \ref{section:framework}, for all contracts $c\in\mathcal{C}$. The relative errors $e_{\text{rel}}(c) =\tfrac{P-\hat{P}}{P}$ of premium values $P\subset c$ and their neural estimate $\hat{P}$ are presented in Figure \ref{fig:errors:relative}. Vertical, dashed lines indicate the $q_\alpha$ quantile for $\alpha\in[0.005, 0.995]$.\\
\begin{table}[bhp]
    \centering
    \input{Plots/male_error_rel_stats}
    \caption{Quantiles $q_\alpha$ of relative errors $e_{\text{rel}}(c)$ [in \%] for $c\in\mathcal{C}$, separately for $\hat{\pi}$ with male and female baseline $\hat{\pi}_{\text{base}}$.}
    \label{tab:rel_errors}
\end{table}

In Figure \ref{fig:errors:relative}, we observe that the relative errors $e_{\text{rel}}(c)$ are approximately symmetrically centered around zero. This holds for both gender specific baselines $\hat{\pi}_{\text{base}}$. Hence, the chosen setting provides an unbiased estimator that is stable for practical initialisations. Comparing the two baselines, we observe that a female baseline $\hat{\pi}$ provides an estimator $\hat{\pi}$ with less pronounced outliers. Overall, the results for both baselines are comparable and $99\%$ of all relative errors $e_{\text{rel}}(c)$ falls approximately in the range of $\pm 5\%$. Given the little information available, in combination with the large number of potentially inhomogeneous probabilities to estimate and the comparably few data points ($N=10,000$) for training, this provides a surprisingly accurate result. Quantitative results for a wider range of $\alpha$-values are provided in Table \ref{tab:rel_errors}. Median relative errors of $-0.01\%$ (male) and $-0.18\%$ (female) confirm, if any, a very low bias of the results. Further, for both baselines $80\%$ of all errors lie within a range of $5\%$, with a slight skew to the negative side. In comparison, we find no consistent difference between the two baselines, other than the more compact range of the female baseline\footnote{Interestingly, the reduced magnitude of outliers for the female baseline is consistent across multiple results throughout our hyperparameter tuning. It is obvious that the male baseline is closer to the final, calibrated transition probabilities than the female baseline, recall e.g. Figure \ref{fig:implied:mortality}. Therefore, we suspect that a baseline that is too close to the final result might introduce a local bias that is hard to reverse during training. }. \\

\begin{figure}[hbt]
    \centering
    \includegraphics[width=.95\textwidth]{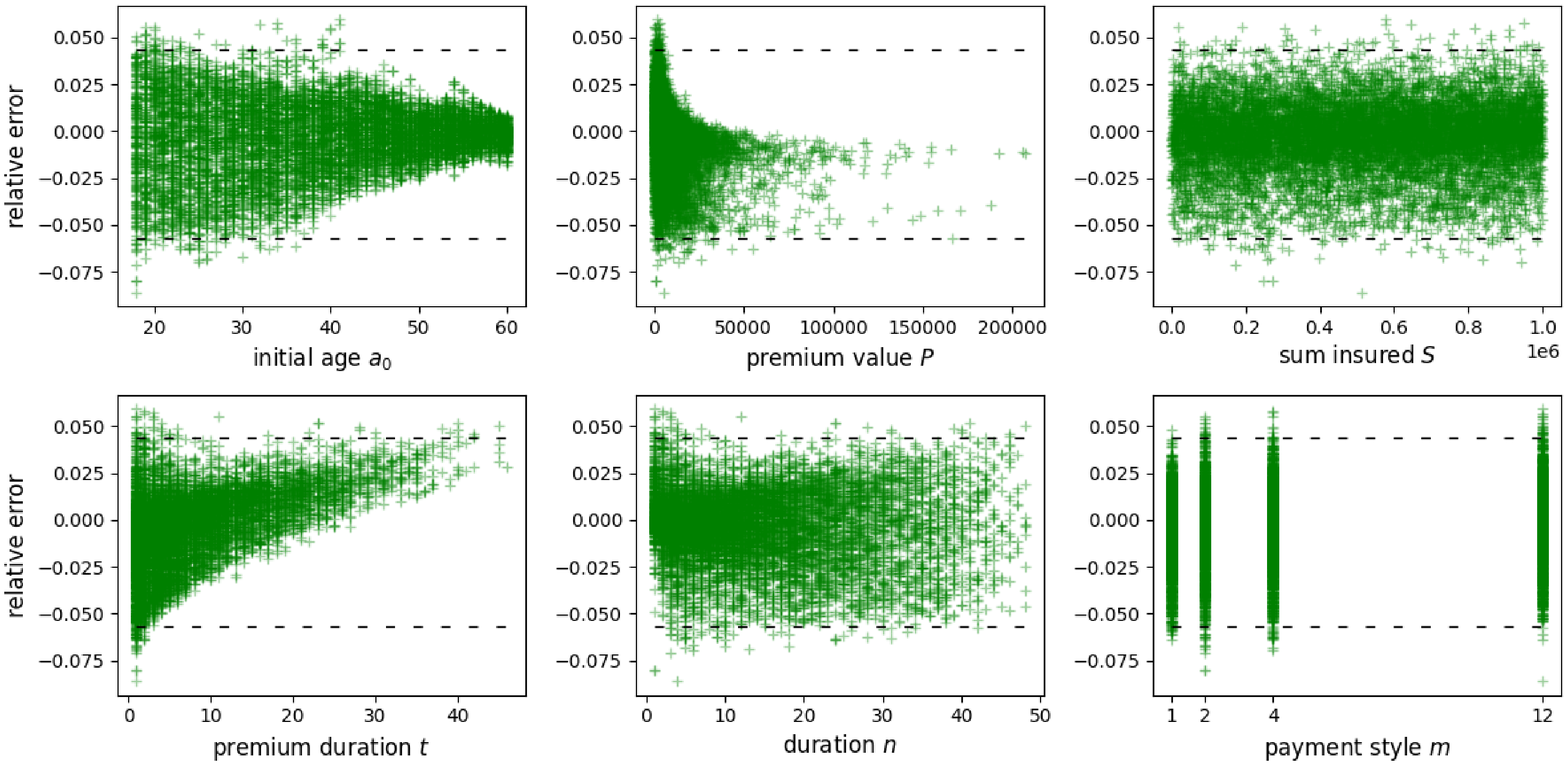}
    \caption{Decomposition of relative error in Figure \ref{fig:errors:relative:female}.}
    \label{fig:errors:relative:female:decomposition}
\end{figure}

As a last step, we investigate the sources for increased relative errors $e_{\text{rel}}(c)$ and sort them by contract features such as the initial ago $a_0$, the premium value $P$, the sum insured $S$, the premium duration $t$ and the payment style $m$. The decomposition based on female baseline $\hat{\pi}$ is provided in Figure \ref{fig:errors:relative:female:decomposition}. The male equivalent contains analogous information and can be found in the Appendix, see Figure \ref{fig:errors:relative:male:decomposition}. \\ 
In Figure \ref{fig:errors:relative:female:decomposition}, we observe that the main explanatory factors for relative errors are the premium value and the duration of premium payments. It is of little surprise that low premium values tend to have higher relative errors, simply because the denominator is reduced. The duration of premium payments, however, shows that short durations exhibit a high variance of  $e_{\text{rel}}(c)$ and a trend where the relative error increases with increasing duration. We explain the later trend by the accumulation of prevalent, minor errors of single-step transition probabilities $\hat{\pi}^{k}$ over an increasing number of iterations. The comparably high variance for low premium durations indicates that our hyperparameters and assumptions, such as the way the administrative charges are included in the premium value, might not be constant for all durations or may differ from our assumption. Recall, that the estimate $\hat{P}$ depends on all our assumptions while the true values $P$ implicitly contains all true, latent hyperparameters, such as the structure of cash flows. Therefor, we can view the decomposition in Figure \ref{fig:errors:relative:female:decomposition} as a diagnostic tool for our original assumptions. The remaining features do not seem to have a systematic effect on $e_{\text{rel}}(c)$. For the initial age, we notice an increased variance for lower initial ages which can easily be explained by a correlation of the initial age and the duration of premium payments. 

%% file: Plots/male_error_rel_stats.tex
\begin{tabular}{l|rrrrrrrrr}
$\alpha$ &  0.000 &  0.005 &  0.100 &  0.250 &  0.500 &  0.750 &  0.900 &  0.995 &  1.000 \\
\midrule
male $q_\alpha$ [\%] & -10.46 &  -5.54 &  -2.59 &  -1.12 &  -0.01 &   1.01 &   2.15 &   5.34 &   9.91 \\
female $q_\alpha$ [\%] &  -8.62 &  -5.69 &  -2.98 &  -1.38 &  -0.18 &   0.86 &   1.91 &   4.38 &   5.96 \\
\bottomrule
\end{tabular}

%% file: Sections/Conclusion.tex
\section{Conclusion} \label{section:conclusion}
In the present work, we introduce a method that can extract inhomogeneous Markov transition probabilities from a portfolio of insurance contracts. Each contract contains an actuarial quantity, such as a premium value, which constitutes a highly lossy compression of the information about the Markov dynamics. No histories of state transitions are available, making the dynamics of the underlying Markov process latent. Instead, we introduce prior information into the setting by assuming a specific structure for state and time-dependent cash flows $CF_{ij}^{(k)}(c)$ that matches the type of contract $c$. In our objective, a custom loss function combines one-step transition probabilities $\hat{\pi}^{(k)}_{ij}(c)$ and cash flows $CF_{ij}^{(k)}(c)$. This allows explicit access to one-step transition probabilities  which can be inspected and validated by backtesting a true, actuarial quantity in contract $c$ with a corresponding estimate. \\
In the numerical analysis, we employ an architecture of a deep, recurrent neural network. Practical, actuarial aspects such as duration of a contract and its premium payments are, by design, respected exactly and explainably in the sequence length of cash flows. We test the approach for a realistic data set of $10,000$ German term life insurance contracts. The number of probabilities to be estimated exceeds the number of data points. However, an economic model validation shows highly accurate results. With a probability of $99\%$, premium values computed with estimated transition probabilities deviate from the true quantities at most by about $5.7\%$. The few outliers with relative errors up to $10\%$ can be explained primarily by low, absolute valued premium values. Overall, the relative errors are approximately symmetric and densely centered around the true value, thereby, numerically affirming the correctness of our method. The results are robust for two different types of initialization.  Further, we are able to extract important information about different profiles of policyholders in the data, such as uncovering an unisex tariff and quantifying the risk spread induced by smoking. From an algorithmic perspective and in line with existing literature, we find empirical evidence for the benefit of transfer learning and deep, overparametrized neural networks for learning our objective. \\

Future research should aim at applying this method to alternative types of contracts and for larger data sets $\mathcal{D}$. Since our approach highly depends on the choice of hyperparameters, such as the cost structure or the actuarial interest rate, we also encourage work that can provide reliable estimates of these quantities. Lastly, it would be interesting to explore a transfer of techniques from neural style transfer in \cite{Gatys.2015}, where a generated image is regularized with respect to a reference, baseline image, to our setting, interpreting the reference image as a reference mortality table. 

%% file: Sections/Appendix.tex
\appendix
\section{Appendix}



\begin{figure}[H]
    \centering
    \includegraphics[width=.85\textwidth]{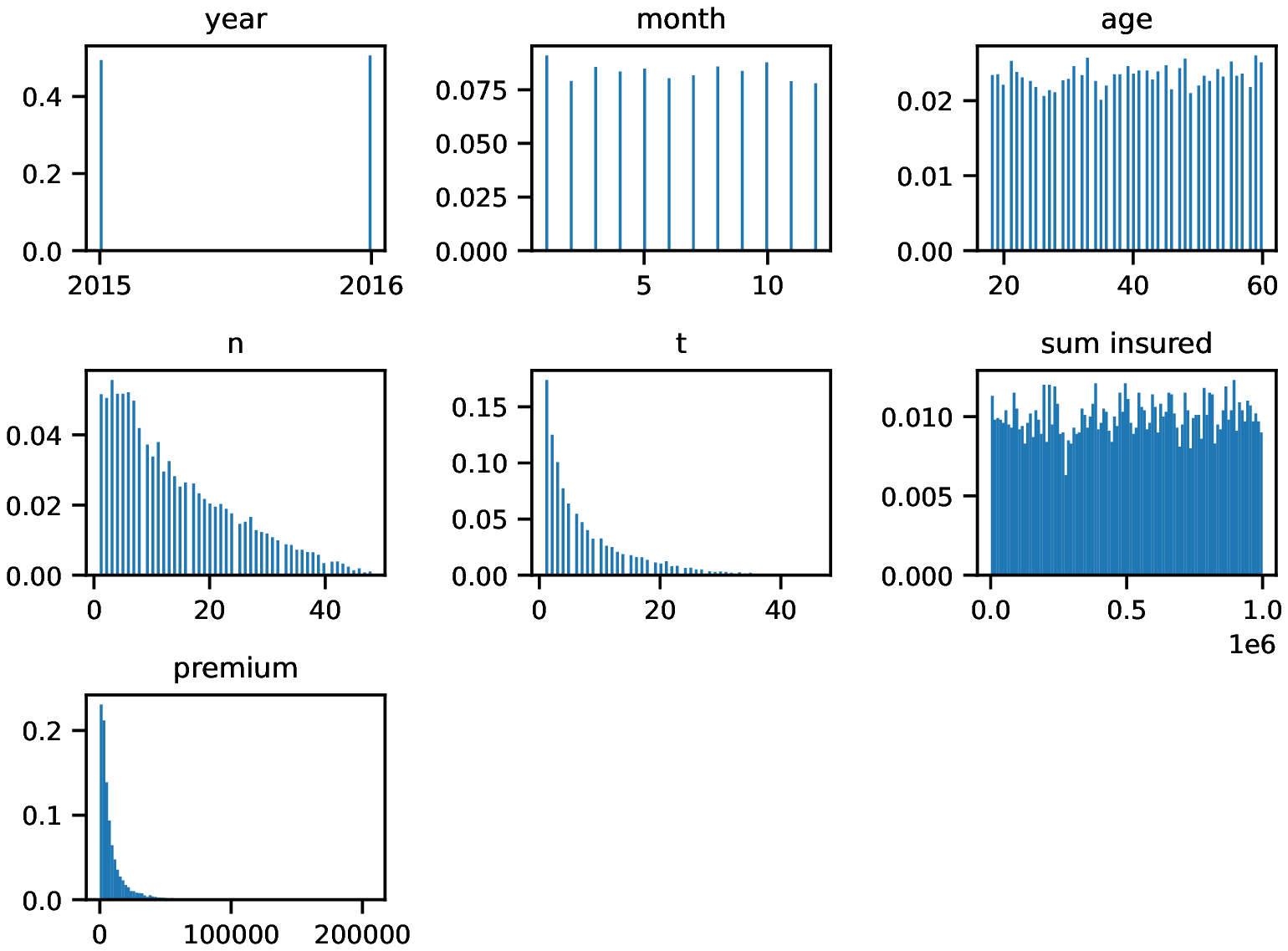}
    \caption{Marginal distributions of the data.}
    \label{fig:appendix:data:marginal:dist}
\end{figure}

\begin{figure}[H]
    \centering
    \begin{subfigure}{.49\textwidth}
        \centering
        \includegraphics[width=\textwidth]{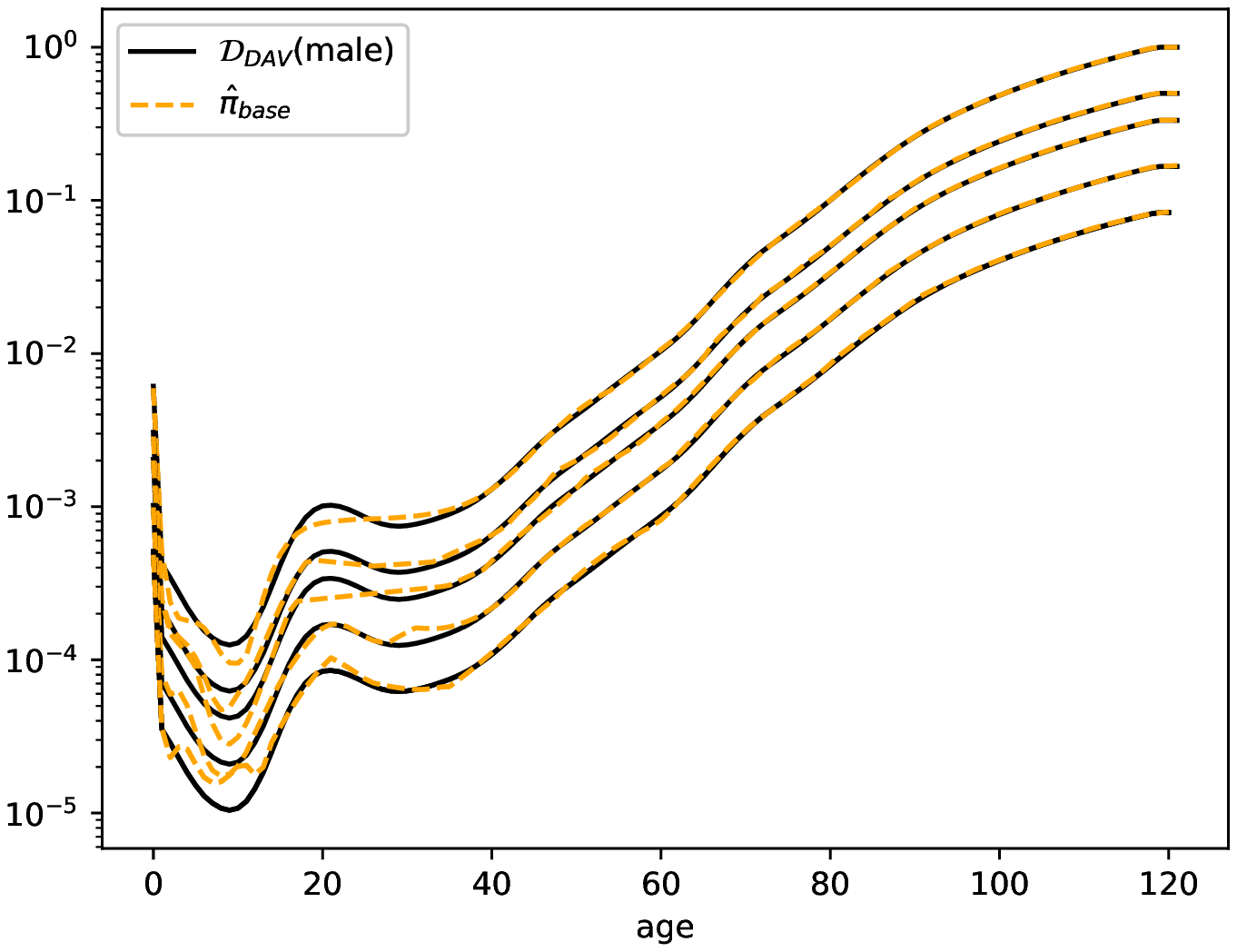}
    \end{subfigure}
    \begin{subfigure}{.49\textwidth}
        \centering
        \includegraphics[width=\textwidth]{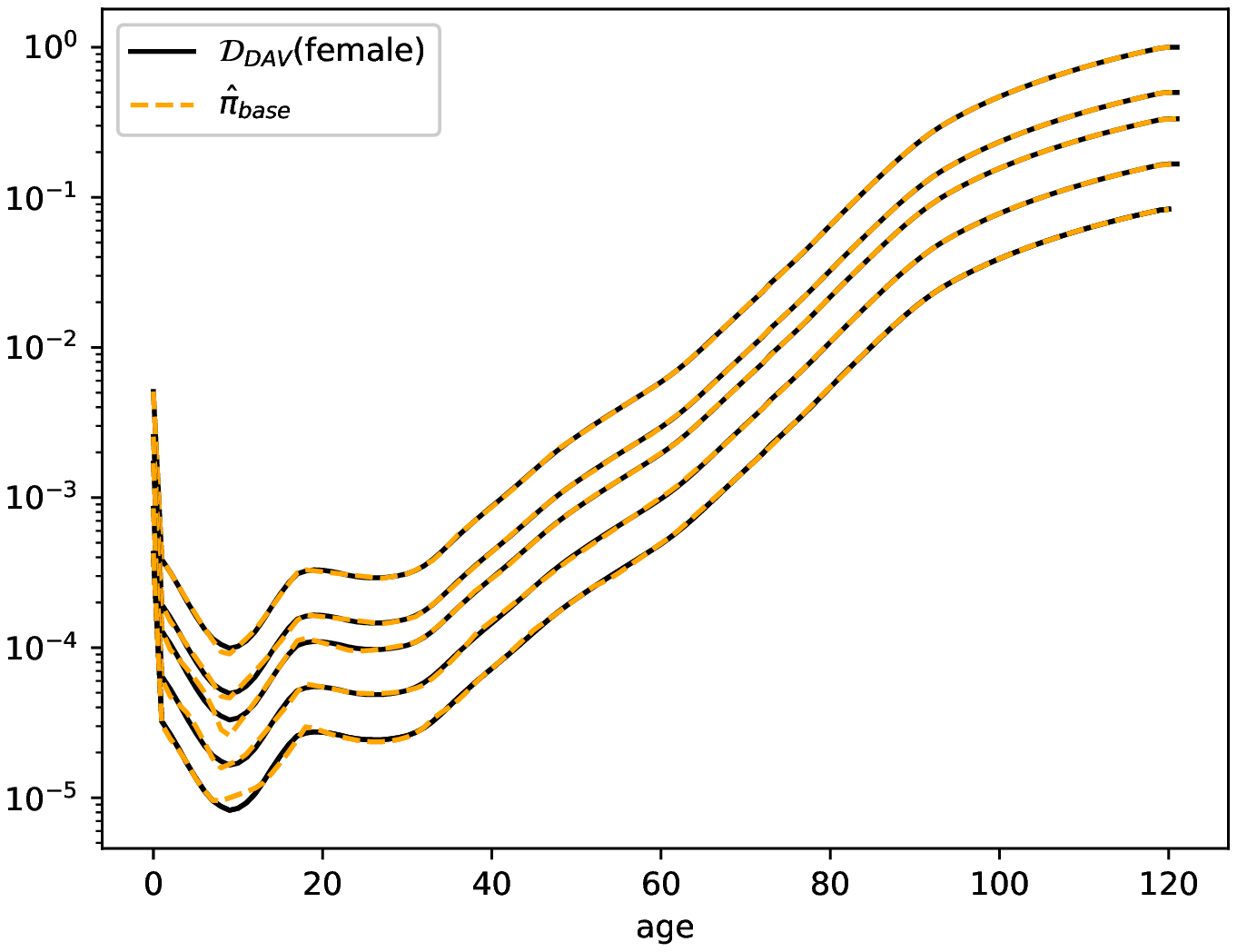}
    \end{subfigure}
    \caption{Death probabilities of the baseline model $\hat{\pi}_{base}$ in comparison to the DAV 2008T table for the gender male (right) and female (left). The parallel lines correspond to the five different payment styles $m\in\lbrace \tfrac{1}{12}, \tfrac{1}{6}, \tfrac{1}{4}, \tfrac{1}{2}, 1 \rbrace$, where a higher value of $m$ corresponds to a higher death probabilitiy. }
    \label{fig:baseline:fit}
\end{figure}

\begin{figure}[H]
    \centering
    \begin{subfigure}{\textwidth}
        \centering
        \includegraphics[width=0.9\textwidth]{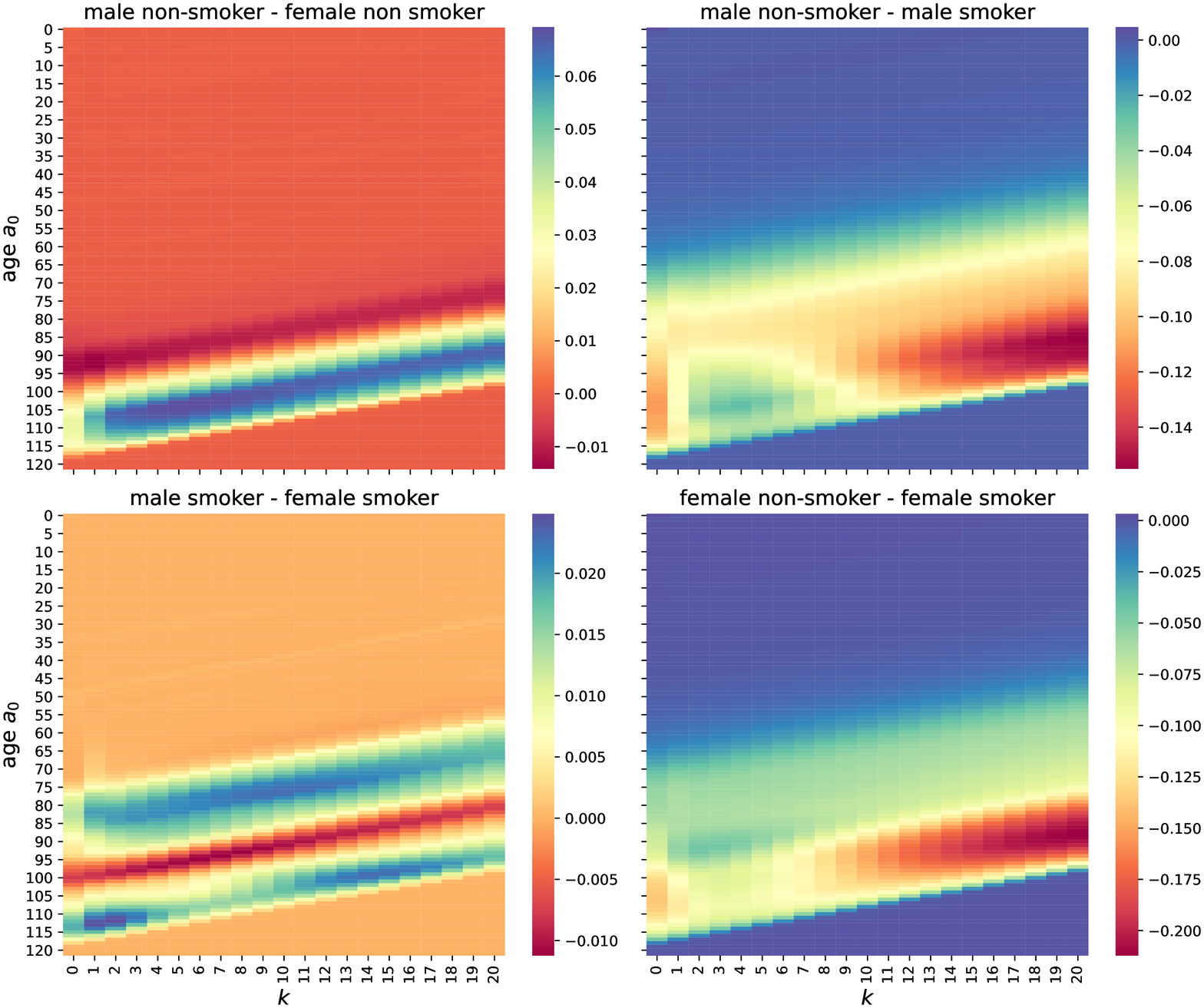}
        \subcaption{$\hat{\pi}_{\text{base}}$ calibrated for DAV2008T male, payment style $m=1$. }
        \label{fig:mortality:heatmap:male}
    \end{subfigure}
    \caption{Information of Figure \ref{fig:mortality:heatmap:male:zoom} for a wider range of initial ages $a_0$.}
\end{figure}
\begin{figure}[H] \ContinuedFloat
    \begin{subfigure}{\textwidth}
    \centering
        \includegraphics[width=0.9\textwidth]{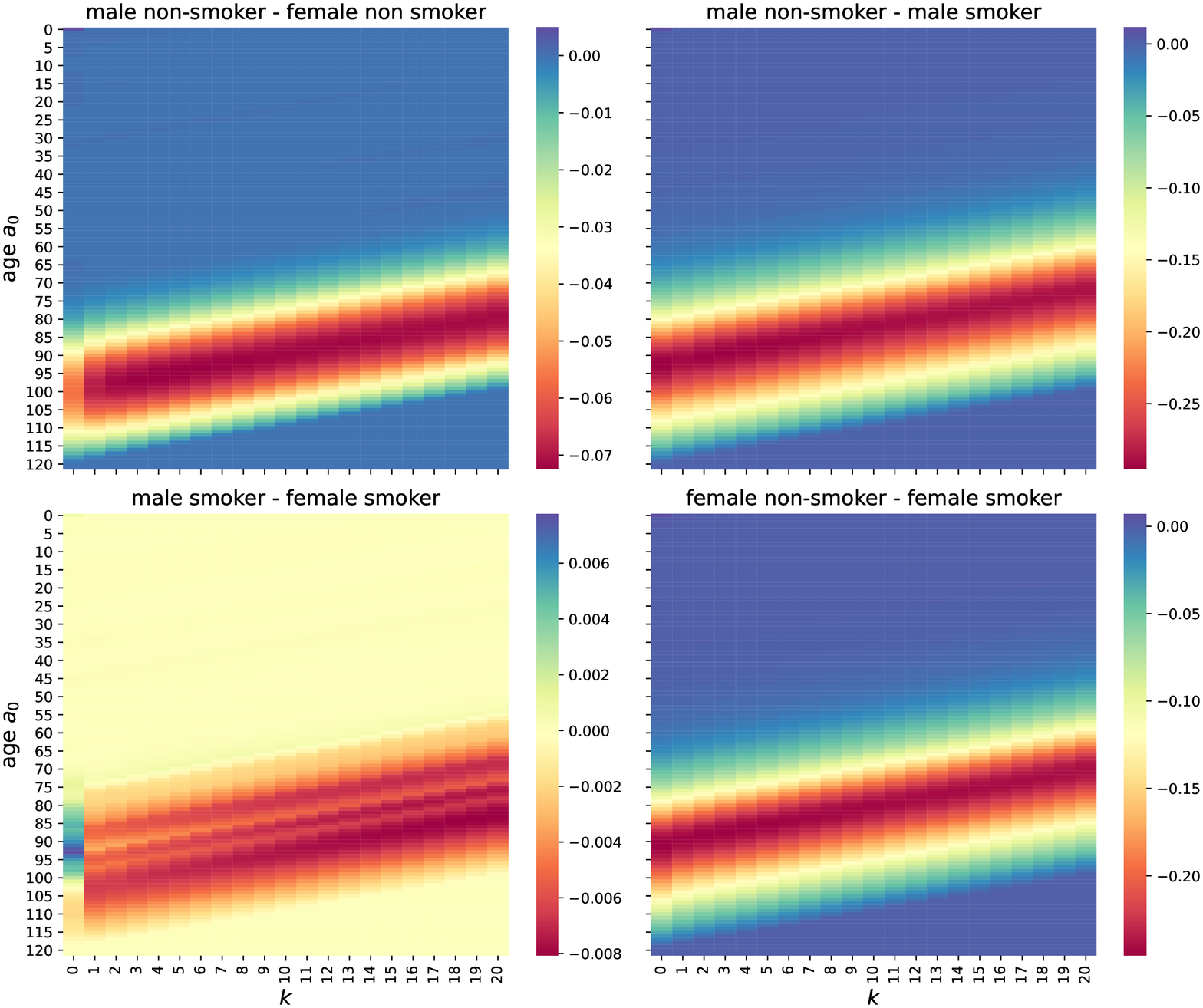}
        \subcaption{$\hat{\pi}_{\text{base}}$ calibrated for DAV2008T female, payment style $m=1$. }
        \label{fig:mortality:heatmap:female}
    \end{subfigure}
    \captionsetup{list=off,format=cont}
    \caption{Information of Figure \ref{fig:mortality:heatmap:female:zoom} for a wider range of initial ages $a_0$.}
    \label{fig:mortality:heatmap:full}
\end{figure}

\begin{figure}[H]
    \centering
    \includegraphics[width=0.9\textwidth]{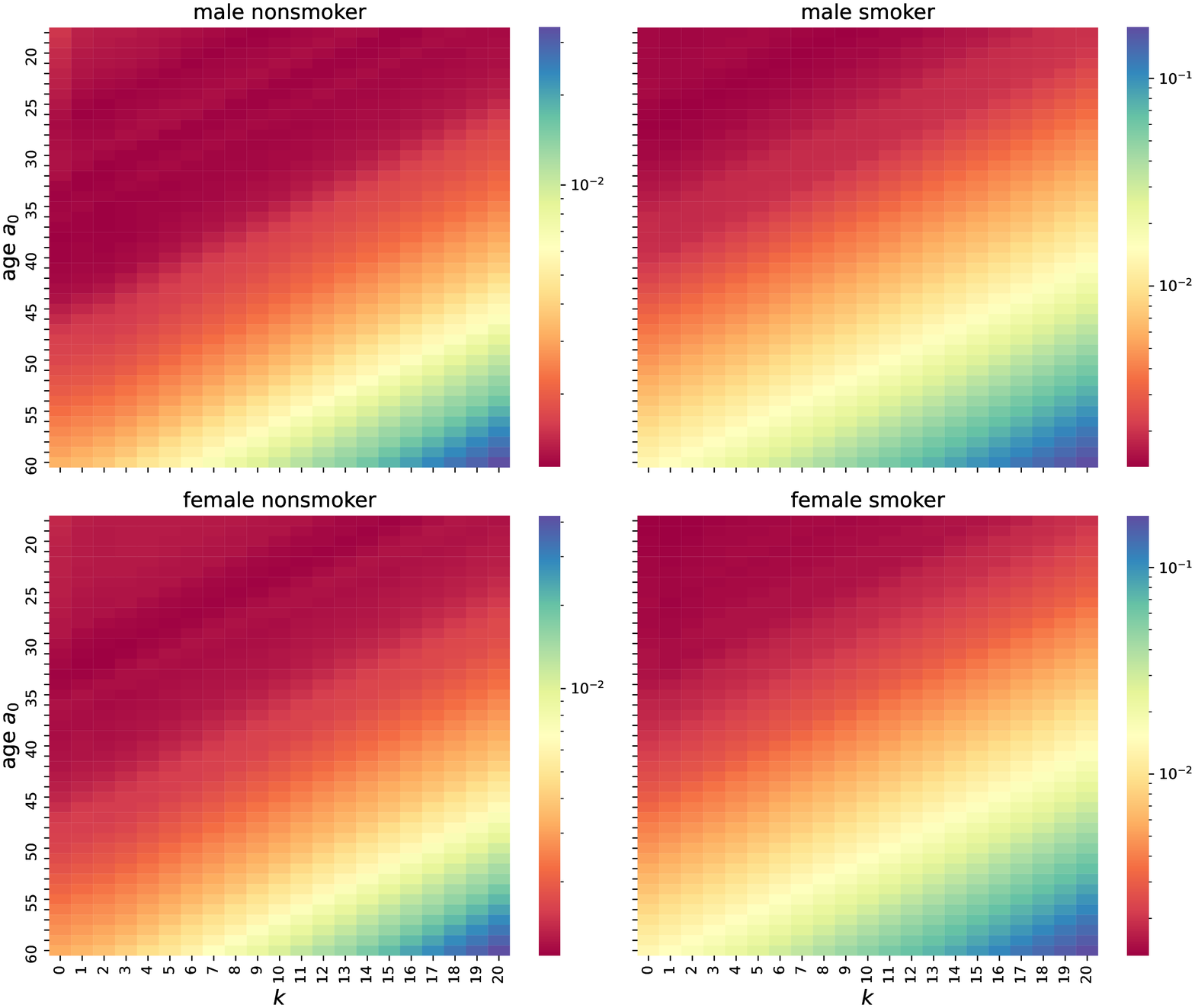}
    \caption{Transition probabilities $\hat{\pi}_{01}(c)$ for different combinations of gender and smoker-status. The payment style is fixed to $m=1$. The underlying baseline $\hat{\pi}_{\text{base}}$ was calibrated on DAV2008T female.}
    \label{fig:mortality:homogeneity:female}
\end{figure}

\begin{figure}[htb]
    \centering
    \includegraphics[width=.95\textwidth]{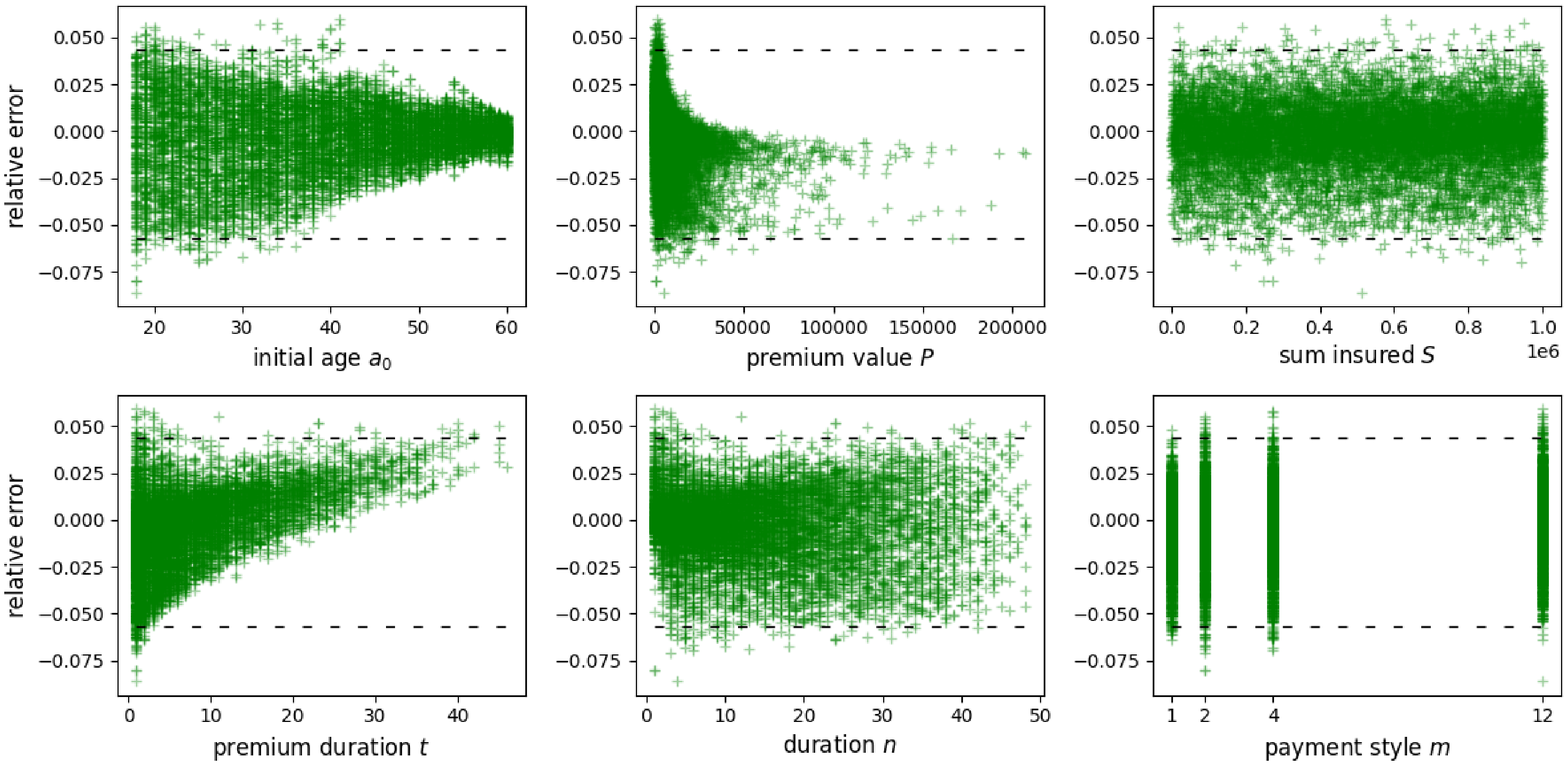}
    \caption{Deomposition of relative error in Figure \ref{fig:errors:relative:male}.}
    \label{fig:errors:relative:male:decomposition}
\end{figure}
